\documentclass[conference]{IEEEtran}
\usepackage{cite}
\usepackage{amsmath,amssymb}
\usepackage{graphicx}
\usepackage{booktabs}
\usepackage{multirow}
\usepackage{makecell}
\usepackage{array}
\usepackage[table]{xcolor}
\usepackage{pifont}
\usepackage{enumitem}
\usepackage{placeins}
\usepackage[hidelinks]{hyperref}
\usepackage{tikz}
\usetikzlibrary{arrows.meta,calc}
\definecolor{txtblue}{HTML}{2B6CB0}      
\definecolor{oursorange}{HTML}{DD6B20}   
\definecolor{stdgray}{HTML}{7A8594}      
\definecolor{goodgreen}{HTML}{2F855A}    
\definecolor{badred}{HTML}{C53030}       

\definecolor{sigred}{HTML}{B00000}
\definecolor{dimgray}{HTML}{6A6A6A}
\definecolor{oursbg}{HTML}{FCF1E7}   
\definecolor{gainbg}{HTML}{E3F1E5}   
\definecolor{lossbg}{HTML}{F7E1E1}   
\definecolor{avgbg}{HTML}{F4F4F4}    
\newcommand{\ci}[2]{\,{\scriptsize[#1,\,#2]}}          
\newcommand{\sig}{\textsuperscript{*}}                   
\newcommand{\cit}[2]{{\scriptsize\color{dimgray}[#1,\,#2]}} 
\newcommand{\gate}{|\!\tanh g|}
\newcommand{\tok}{\texttt{<kg\_node>}}
\newcommand{\secref}[1]{Sec.~\ref{#1}}
\newcommand{\appref}[1]{App.~\ref{#1}}
\setlist{leftmargin=*, itemsep=1pt, topsep=2pt}
\newif\ifshowpending\showpendingtrue

\begin{document}

\title{Do Embodied VLMs Actually Use Their Scene-Graph Memory?}

\author{\IEEEauthorblockN{Shubh Garg\textsuperscript{1,*}\quad
Divyansh Pandey\textsuperscript{1,2,*}\quad
Vedant Khandelwal\textsuperscript{3}\quad
Niyati Rawal\textsuperscript{1,4}\quad
Amit Sheth\textsuperscript{1}}
\IEEEauthorblockA{\textsuperscript{1}Indian AI Research Organisation (IAIRO), GIFT City, Gujarat\quad
\textsuperscript{2}UPAI Labs\\
\textsuperscript{3}AI Institute, UofSC (AIISC), 1112 Greene St., Columbia\quad
\textsuperscript{4}BITS Pilani\quad
\textsuperscript{*}Equal contribution\\
{\footnotesize\texttt{\{shubh.garg@iairo.ai, shubhgarg265@gmail.com\}, \{divyanshpandey0108@gmail.com, divyansh@upailabs.org\},}}\\
{\footnotesize\texttt{kh.vedant02@gmail.com, niyati.rawal@iairo.ai, \{amit.sheth@iairo.ai, amitpsheth@gmail.com\}}}}}

\maketitle

\begin{abstract}
Embodied agents that answer questions about their route need a memory of what they saw, and scene graphs are a natural form for it. A vision-language model (VLM) can receive such a graph as text or through a graph encoder, but accuracy alone cannot show whether the graph  holds the episode's facts, reaches the model, or is actually relied on. In this diagnostic study, we introduce a graph-swap test for the last of these: each episode is given another episode's graph, and we count how many answers change. On an embodied question-answering benchmark with routes through two to seven rooms, we use it to compare ways of delivering scene-graph memory to a fine-tuned VLM. As text, the memory is used strongly, adding 30 points on route questions. As a graph, it is not, and we trace why: the encoder preserves the information, but standard pooled and late fusion discard it. Giving each node its own input token recovers part of it, a gain of 8.3 points on route questions that the swap test confirms comes from the graph. The test also exposes a shortcut that accuracy hides: when the same facts appear in the text, the model ignores its graph. The test thus offers a cheap check of whether a VLM uses its scene-graph memory, and our analysis points to how that memory should be delivered.
\end{abstract}

\section{Introduction}\label{sec:intro}
Embodied navigation is partially observed. An agent moving through a house sees only part of it at any moment, so to answer questions later about where it has been, it needs an episodic memory of what it saw. Scene graphs are a natural spatial representation for that memory. They store rooms, objects and relations, a topological and semantic map that recent systems build online for navigation and planning~\cite{hughes2022hydra,gu2024conceptgraphs,werby2024hovsg,rana2023sayplan,yin2024sgnav}. Once the graph is built, a vision-language model (VLM) can be grounded in it in two ways. The graph can be written into the prompt as text~\cite{fatemi2024talk}, a loose coupling that leaves the model to parse it. Or a graph neural network can encode it, and the node embeddings enter the model as extra input vectors~\cite{perozzi2024graphtoken,he2024gretriever,chen2024llaga,tang2024graphgpt}. This tighter coupling keeps the graph's structure in the model's input, but neither coupling guarantees that the model uses the graph when it answers.

Whether a graph memory helps depends on three separate things. \emph{Quality}: does the graph hold the facts of the episode? \emph{Access}: does its content reach the model? \emph{Dependence}: do the model's answers change when the episode-specific content changes? Accuracy mixes all three. A fine-tuned model can score well from text, from answer priors, or from the mere presence of a memory field, while its graph goes unread. Work on feeding graphs to language models compares ways of delivering a graph (\secref{sec:related}), but delivery and task accuracy do not show that the model relies on the graph of the episode at hand. We study all three properties; dependence is the central one.

\begin{figure*}[t]
\centering
\includegraphics[width=\textwidth]{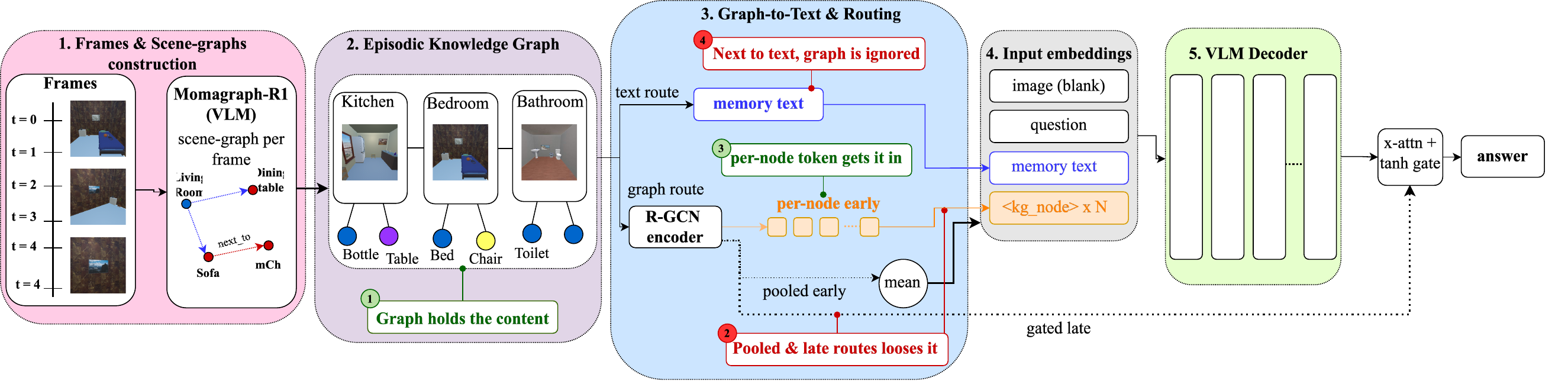}
\caption{Our scene-graph memory and three ways to give it to a VLM, with where its information is lost. Our pipeline (left) turns the agent's frames into per-frame scene graphs and merges them into one episode graph, which reaches the model as memory text (blue), through a graph encoder (R-GCN), or both. The standard graph route (dashed) combines a mean-pooled vector added at the input and a gated cross-attention after the last layer. Our per-node route writes one token per node into the prompt. Numbered tags: the four findings of \secref{sec:intro}, placed on the part of the pipeline each concerns.}
\label{fig:overview}
\end{figure*}

We work on an embodied question-answering benchmark whose routes pass through two to seven rooms (\secref{sec:bench}). For quality, we propose a validated pipeline that builds the memory (\secref{sec:sg}): per-frame scene graphs from MomaGraph-R1~\cite{ju2026momagraph} on $512{\times}512$ frames with a closed object vocabulary, per-type correction rules fit on validation data, and rooms segmented at the agent's door crossings, merged into one episode graph and scored against the simulator's records. For access, we hold the graphs, backbone, training data, schedule and evaluation fixed and vary the route by which the memory enters the VLM (\secref{sec:routes}, Fig.~\ref{fig:overview}); the graph routes differ in their fusion modules, which are trained with the model. For dependence, a graph-swap test scores every episode with another episode's graph and compares the answers that change with re-evaluation noise (\secref{sec:swap}). Three questions organise the paper. \textbf{RQ1}: how does text memory compare with graph routes under this control (\secref{sec:acc}--\ref{sec:pernode})? \textbf{RQ2}: does a per-node graph route make the VLM depend on episode-specific graph content (\secref{sec:swap})? \textbf{RQ3}: how do perception quality, question type and route length limit the gains (\secref{sec:failures})? Four findings follow.
\begin{enumerate}
\item \textbf{The graphs hold the content.} Written as text, they lift accuracy from 64.8\% to 75.9\% (two seeds), with a gain of 30.1 points on questions about the route.
\item \textbf{The standard graph route loses it.} A mean-pooled vector at the input together with a gated cross-attention after the last layer leaves a graph-only model within noise of no memory (+1.6 points).
\item \textbf{One token per node gets part of it in.} A graph-only model now gains 8.3 points on route questions and 18.1 on ``how many rooms did the robot visit'', and another episode's graph costs it 6.5 points on route questions.
\item \textbf{Next to text, the graph is ignored.} With memory text present, swapping the graph changes 12 of 800 answers, as many as re-running the evaluation (9 to 12), while swapping the text costs 16.4 points. We detect no dependence on the graph in this condition.
\end{enumerate}
These findings hold for one backbone, one benchmark, our scene graphs and the routes we tested.
The paper is a diagnostic study. Text memory is the reference a graph route must beat, and the per-node route's accuracy gain over the standard route is small and within noise (67.1\% against 65.9\%); what it adds is a graph dependence that the swap test detects.

\textbf{Contributions.} Our central contribution is (i) a graph-swap evaluation protocol that tells ``the graph reaches the model'' apart from ``the model needs the graph'', calibrated against re-evaluation noise. It rests on (ii) a pipeline that builds an episodic scene-graph memory from an agent's frames, with its output scored against simulator records, and (iii) a controlled comparison of memory routes into a VLM (text, the standard pooled-input and late cross-attention route, and per-node tokens) on one set of scene graphs. The protocol yields (iv) a per-node input route whose graph-only model measurably depends on its graph, and (v) a failure analysis of where scene-graph memory stops helping. All logs, predictions and code are released with the paper (\appref{app:logs}).

\textbf{Terms.} The \emph{episode graph}, or knowledge graph (KG), is the memory of one episode, merged from per-frame scene graphs. The graph encoder turns it into node embeddings. A \emph{memory route} is how the memory enters the VLM: as text, through the standard graph route, or through per-node tokens. Text-only, graph-only and text + graph name which routes a model reads. A model \emph{uses} a memory when its answers depend on the episode-specific content, which is what the swap test measures.

\section{Related Work}\label{sec:related}
\textbf{Scene graphs and maps as memory.} 3D scene graphs organise a space into rooms and objects, a map that is both topological and semantic~\cite{hughes2022hydra,maggio2024clio}. With open-vocabulary features, graphs and maps are used for navigation and planning~\cite{gu2024conceptgraphs,werby2024hovsg,huang2023vlmaps,rana2023sayplan,yin2024sgnav}, and long-horizon robot memories can be queried in language~\cite{anwar2025remembr}. In embodied question answering, an agent answers from what it observed~\cite{das2018eqa,majumdar2024openeqa,ren2024explore}. Recent systems give the model its memory as serialised scene graphs~\cite{saxena2025grapheqa}, scene graphs it can edit~\cite{ali2025graphpad}, selected images~\cite{yang2025mem} or memory features~\cite{hu2025mem}, and multimodal models still struggle to recall space from video~\cite{yang2025thinking}. These systems take one step for granted: how the graph's content reaches the model. We study that step.

\textbf{Feeding graphs to language models.} Graphs can be written as text, and the wording matters~\cite{fatemi2024talk,wang2023nlgraph}. A graph encoder can instead produce input vectors: one pooled vector~\cite{perozzi2024graphtoken,tian2024gnp}, one projected embedding per node~\cite{chen2024llaga,tang2024graphgpt}, or a graph vector next to retrieved text~\cite{he2024gretriever}. The projector design follows vision-language models~\cite{liu2023llava}, and 3D language models give each object its own token~\cite{hong20233dllm,huang2024chatscene}. Fusion can also happen inside the network, layer by layer~\cite{zhang2022greaselm} or through a cross-attention gated at zero~\cite{alayrac2022flamingo}. These works report that such vectors can match or beat text on graph tasks. None, to our knowledge, compares delivery routes for an episodic scene-graph memory with the encoder held fixed, or tests whether the model needs the graph when the same facts are in the text.

\textbf{Shortcuts and ignored inputs.} Multimodal networks often learn mainly from the input that is easiest to fit and under-use the rest~\cite{wang2020makes,wu2022greedy}. Question-answering models answer from language priors~\cite{agrawal2018vqacp}, and single-input baselines are strong on embodied tasks~\cite{thomason2019shifting}. Benchmarks can reward such shortcuts~\cite{geirhos2020shortcut}. Dropping one input during training is a standard remedy~\cite{neverova2016moddrop}. These studies measure under-use of a whole input through accuracy. We ask a narrower question with a swap test: does the model need the episode-specific content of the memory it is given?

\section{Benchmark}\label{sec:bench}
Each episode is a route in a ProcTHOR-10k house~\cite{deitke2022procthor} in AI2-THOR~\cite{kolve2017ai2thor}. The agent walks from its start to a target object in another room, through two to seven rooms of four types (bathroom, bedroom, kitchen, living room). Each step stores a $256{\times}256$ RGB frame; episodes average 41.7 frames (10 to 116). Episodes are balanced over one, two, three and four or more door crossings (strata S1 to S4). At the end, the agent answers four multiple-choice questions (two to four options; chance 33\%), generated from the simulator's record of what was visible. Eight question types fall into two groups (Table~\ref{tab:qtypes}). \emph{Route} questions ask when things were seen: the order in which objects first appeared, whether an object was still in view at the end, and how many rooms were visited (291 test questions). \emph{Set} questions ask what was where: in how many rooms an object appeared, its room, which option was seen in a given room, and object counts (509). Questions are worded about the episode, but each gold answer is computed from six frames sampled along the route (the question strip); the memory holds every frame before the question. Splits are house-disjoint: 5,000 training, 500 validation and 200 test episodes from 1,359, 151 and 53 houses, with 800 test questions. \appref{app:data} gives statistics and sample frames.

\begin{table*}[t]
\centering
\caption{Question types, gold answers and examples. Options are in brackets. Every gold answer is computed from the visible-object records and room labels of a strip of six sampled frames.}
\label{tab:qtypes}
\footnotesize
\setlength{\tabcolsep}{4pt}
\renewcommand{\arraystretch}{1.08}
\begin{tabular}{@{}>{\raggedright\arraybackslash}p{2.2cm}r>{\raggedright\arraybackslash}p{5.6cm}>{\raggedright\arraybackslash}p{8.4cm}@{}}
\toprule
Type & $n$ & Gold answer & Example test question\\
\midrule
\multicolumn{4}{@{}l}{\emph{Route questions: when things were seen} (291)}\\
Order & 73 & whether type $X$ is first seen before or after type $Y$ & Did the robot first see a sink before or after it first saw a houseplant? [before, after]\\
Presence change & 102 & whether a type seen in an earlier strip frame is absent from the strip's last frame & Was a statue visible at some point but then gone from view later? [yes, no]\\
Rooms visited & 116 & number of distinct room labels among the strip frames & How many different rooms did the robot enter during its run? [2, 3, 4, 5]\\
\midrule
\multicolumn{4}{@{}l}{\emph{Set questions: what was where} (509)}\\
Persistence & 131 & number of distinct room labels whose frames show the type & In how many different rooms did the robot see a painting? [2, 3, 4]\\
Room ID & 80 & room label of the frames that show the type & In which room did the robot see a bed? [LivingRoom, Bedroom, Bathroom]\\
Existence & 113 & the option type seen in frames of the named room & Which of these did the robot see in the Bathroom? [desklamp, plunger, statue, pencil]\\
Counting & 65 & distinct instance IDs of the type in frames of the named room & How many distinct chair objects did the robot see in the Kitchen? [1, 2, 3]\\
Global count & 120 & distinct instance IDs of a category (e.g., storage units) over the strip & In total, how many distinct electronic objects did the robot see anywhere in the house? [3, 4, 5]\\
\bottomrule
\end{tabular}
\end{table*}

\section{Scene-Graph Construction}\label{sec:sg}
The questions need four kinds of fact from the memory: which objects were seen, in which room, when each was first and last seen, and how objects were placed relative to each other. The pipeline is built to keep these.

\textbf{Per-frame graphs.} MomaGraph-R1~\cite{ju2026momagraph}, a Qwen2.5-VL-7B model trained with reinforcement learning to output task-oriented scene graphs, describes each frame as a room label, a list of objects and spatial relations between objects from a fixed set (left of, right of, in front of, behind, higher, lower, close, far, touching). Our pipeline turns these per-frame outputs into an episodic memory in four steps . Fig.~\ref{fig:overview} shows how the scene graph memory enters into VLM.

\textbf{(1) Resolution.} Small and distant objects need pixels, so frames are rendered at $512{\times}512$ from the stored agent poses. \textbf{(2) Closed vocabulary.} The prompt (Fig.~\ref{fig:prompts}a) lists the benchmark's object types, asks for small objects and leaves out walls, doors and floors, so that every name is a type the questions can use. \textbf{(3) Per-type rules.} False positives repeat by type. One rule per type, fit greedily on the validation split, keeps the type, drops it or renames it within its family; the fit returns 28 rules. \textbf{(4) Rooms at door crossings.} A room label guessed anew in every frame is unstable, so the frames between two consecutive door crossings, taken from the simulator's record of the route, form one room. A classifier names each room from the objects and per-frame room guesses in it, and a repeated room type is numbered (``Bedroom 2''). Table~\ref{tab:sgquality} scores the resulting graphs against the simulator's records: they find half of the visible object types at precision 65.7, name the room correctly in 72\% of frames and get the number of visited rooms right in 199 of 200 test episodes. \appref{app:sg} gives every detail, including the full rule list.

\begin{table}[t]
\centering
\caption{Quality of our scene graphs on the test split (200 episodes, the 8,135 frames the answering model may use), scored against the simulator's visible-object records. P, R, F1: object types per frame (\%). Frame: frames with the right room label (\%). Count: episodes with the right number of visited rooms (of 200). Edit: edit distance between the remembered and the true room sequence (lower is better). Obj.--room: recall of (object type, room) pairs (\%). Order: pairs of types first seen in the right order (\%).}
\label{tab:sgquality}
\footnotesize
\setlength{\tabcolsep}{3.4pt}
\begin{tabular}{@{}ccc c ccc c cc@{}}
\toprule
\multicolumn{3}{c}{Objects} & & \multicolumn{3}{c}{Rooms} & & \multicolumn{2}{c}{Episode memory}\\
\cmidrule{1-3}\cmidrule{5-7}\cmidrule{9-10}
P & R & F1 & & Frame & Count & Edit & & Obj.--room & Order\\
\midrule
65.7 & 49.2 & 56.3 & & 71.8 & 199 & 1.30 & & 45.5 & 78.4\\
\bottomrule
\end{tabular}
\end{table}

\textbf{Episode graph.} The frames seen so far are merged into one episode graph (the KG of \secref{sec:intro}). Room nodes follow the visit order. An object node is an object type filed under a room, with the first and last frame in which it was seen and the largest number seen in one frame; relations become edges between object nodes. The KG records types, not tracked instances: two chairs in one room are one node with a count of two. It is built only from frames the answering model may use.

\textbf{Perfect scene graphs.} As an upper bound, we also build KGs from the simulator's own records of the visible objects (with stable identities), the rooms and the spatial relations. No deployed agent has them.

\begin{figure}[t]
\centering
\includegraphics[width=\columnwidth]{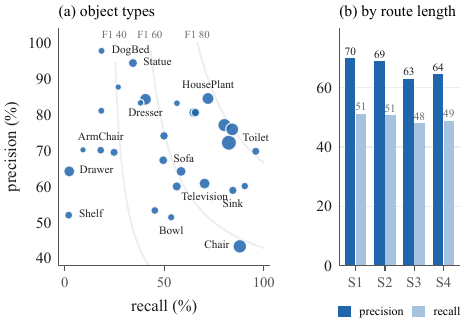}
\caption{What our scene graphs find on the test split (8,135 frames, scored by object type against the simulator's records). (a) Precision and recall of each object type visible in at least 300 frames and kept by the rules; dot area: frames in which the type is visible; gray curves: equal F1. (b) Object precision and recall by route stratum (one to four or more door crossings).}
\label{fig:sgtypes}
\end{figure}

\textbf{What the graphs find.} Recall depends mostly on the object type (Fig.~\ref{fig:sgtypes}a). Large fixtures are found in most frames where they are visible (toilet 96\%, chair 88\%, sink 85\%, dining table 83\%), and the ten most frequent types reach 67.8\% recall together. Storage furniture and its parts are rarely named: drawers and shelves in 2\% of their frames, armchairs in 18\%, boxes in 25\%. The 25 types the rules remove make up 11\% of the visible types and 22\% of the misses, a choice of precision over recall. Chairs alone cause 26\% of the false positives, since the model lists a chair in about twice as many frames as one is visible. Recall barely depends on distance: for the nearest visible instance of a type it stays between 47\% and 51\% from under 1.5\,m to beyond 4.5\,m, and objects seen through a doorway are found as often as those in the agent's room. Per-frame quality drops only slightly on longer routes (Fig.~\ref{fig:sgtypes}b; object F1 59.2 with one door crossing, 54.6 and 55.5 with three or more).

\section{Memory Routes}\label{sec:routes}
\textbf{Arms.} Every model is MomaGraph-R1 (Qwen2.5-VL-7B~\cite{bai2025qwen25vl}) fine-tuned on the benchmark. All models share the training data, schedule, LoRA settings and evaluation; they differ in the memory they read and, for graph models, in the modules that deliver it (Table~\ref{tab:arms}). Arm~A (\emph{no memory}) sees the question and options only. Arm~B (\emph{text}) reads the KG written into the prompt. Arm~C (\emph{graph}) reads it only through a graph encoder, and arm~D (\emph{text + graph}) reads both. Arms C and D are trained once with the standard graph route and once with our per-node route. At answer time the image is blank, so every fact about the episode must come from memory. Fig.~\ref{fig:prompts} shows the prompts.

\begin{table}[t]
\centering
\caption{The arms. All are MomaGraph-R1 fine-tuned with LoRA on the same data and schedule; they differ only in how the episode KG reaches the model. Seeds: training seeds reported.}
\label{tab:arms}
\footnotesize
\setlength{\tabcolsep}{3pt}
\begin{tabular}{@{}l l c l c@{}}
\toprule
Arm & Memory & Text & Graph route & Seeds\\
\midrule
A & No memory & -- & -- & 2\\
B & Text & \ding{51} & -- & 2\\
C & Graph & -- & standard (pooled + late) & 1\\
D & Text + graph & \ding{51} & standard (pooled + late) & 1\\
\rowcolor{oursbg}C & Graph & -- & per-node tokens (ours) & 2\\
\rowcolor{oursbg}D & Text + graph & \ding{51} & per-node tokens (ours) & 2\\
\midrule
B & Text, perfect scene graphs & \ding{51} & -- & 2\\
\bottomrule
\end{tabular}
\end{table}

\begin{figure*}[t]
\centering
\setlength{\fboxsep}{4pt}
\fcolorbox{black!35}{black!2}{\begin{minipage}[t]{0.405\textwidth}
\raggedright\ttfamily\fontsize{5.6}{6.7}\selectfont
{\rmfamily\footnotesize\bfseries (a) Per-frame scene-graph prompt (closed vocabulary)}\\[3pt]
You are an embodied robot perception system. List the objects visible in this image.\\[2pt]
STAGE 1 --- ROOM: scene\_type is the room you are CURRENTLY IN, judged only from the image. Exactly one of: "Bathroom", "Bedroom", "Kitchen", "LivingRoom".\\[2pt]
STAGE 2 --- OBJECTS: list EVERY object you can see, including SMALL objects (cups, books, bottles, boxes, pillows, remote controls, food, drawers, shelves) and objects in the background or through doorways. Use ONLY these exact type names:\\
AlarmClock, Apple, ArmChair, BaseballBat, \dots, WineBottle {\rmfamily\itshape [106 types, App.~\ref{app:prompts}]}\\
If an object matches none of these names, leave it out. Do NOT list walls, floors, doors, windows or hallways. Number duplicates (Chair1, Chair2). Do not invent hidden objects.\\[2pt]
STAGE 3 --- EDGES: spatial relationships between listed objects, using exactly: left\_of, right\_of, in\_front\_of, behind, higher\_than, lower\_than, close, far, touching.\\[2pt]
Output EXACTLY:\\
<SCENE\_GRAPH\_ANALYSIS>\\
One short paragraph naming the room and the objects you see.\\
</SCENE\_GRAPH\_ANALYSIS>\\
<SCENE\_GRAPH\_JSON>\\
\{"scene\_type": "...", "target\_scene": "unknown", "nodes": [\{"object": "Chair1"\}], "edges": [\{"object1": "...", "object2": "...", "spatial\_relationships": ["..."]\}]\}\\
</SCENE\_GRAPH\_JSON>
\end{minipage}}\hfill
\fcolorbox{black!35}{black!2}{\begin{minipage}[t]{0.555\textwidth}
\raggedright\ttfamily\fontsize{5.6}{6.7}\selectfont
{\rmfamily\footnotesize\bfseries (b) Answer prompt, arm B (text memory), a real test episode}\\[3pt]
\#\#\# Task: Navigation has finished.\\
Answer the question using the memory collected during navigation.\\
\#\#\# Question: Did the robot first see a sink before or after it first saw a houseplant?\\
\#\#\# Choices: A: Before B: After\\
\#\#\# Memory:\\
ROOMS: {\color{badred}room\_kitchen}$^{1}$, room\_kitchen\_2, room\_bedroom, room\_bathroom\\
VISITS (frames f0-f42): {\color{badred}room\_kitchen}@f0-f11 -> room\_kitchen\_2@f12-f19 -> room\_bedroom@f20-f31 -> room\_bathroom@f32-f42\\
room\_kitchen: Painting x2[f0-f11], HousePlant[f0-f11], Chair[f3-f8], {\color{badred}Fridge x2[f7-f11]}$^{2}$, {\color{badred}Bed[f0-f2]}$^{3}$, Television[f0-f10], Bedding1[f0-f1], GarbageBag[f10-f11], TVStand[f0], DiningTable[f8], Rock[f9], Oven[f10]\\
room\_kitchen\_2: Painting[f12-f17], Chair x2[f12-f19], Fridge[f12-f16], Bowl[f12-f14], DiningTable[f12-f14], GarbageBag[f12], Television[f12], Rock[f13], {\color{badred}Refrigerator Magnets[f14], Shadow[f17]}$^{4}$\\
room\_bedroom: Chair x2[f20-f29], DiningTable[f22-f29], Bed[f26-f27], {\color{badred}Faucet[f30-f31], Sink[f30-f31], Toilet[f30-f31]}$^{2}$, TVStand[f22], Television[f22], Dresser[f26], Statue[f26], Book[f27], Box[f27]\\
room\_bathroom: GarbageCan[f32-f41], Toilet[f32-f42], Faucet[f32-f34], Sink[f32-f34], GarbageBag[f34], {\color{badred}TiledWall[f40]}$^{4}$, Shadow[f42]\\
RELATIONS: Bed-in\_front\_of-Bedding1; Television-in\_front\_of-TVStand; Bed-left\_of-Painting; Bed-right\_of-HousePlant; Chair-in\_front\_of-DiningTable; \dots {\rmfamily\itshape [24 relations]}\\
\#\#\# Guidance:\\
Select the correct answer from the given options.\\
\#\#\# Answer: {\rmfamily\itshape (gold: B)}\\[5pt]
{\rmfamily\fontsize{6.4}{7.6}\selectfont
\textbf{Errors against the simulator records} (red). Room changes are right; but $^{1}$the first room, a living room, is named Kitchen (so the real kitchen becomes ``kitchen\_2''); $^{2}$objects seen through a doorway are filed under the agent's room (fridge: kitchen; sink, faucet, toilet: bathroom); $^{3}$no bed stands there, and the sofa is never listed; $^{4}$a few out-of-vocabulary names slip through. The order question is still answerable: the houseplant is first seen at f0 and the sink at f30.}
\end{minipage}}\\[3pt]
\fcolorbox{black!35}{black!2}{\begin{minipage}{0.977\textwidth}\raggedright\fontsize{6.4}{7.6}\selectfont
\textbf{(c) What changes between arms.} \textbf{A} (no memory): no \texttt{\#\#\# Memory} block, and the second line reads ``\dots using the visual information collected during navigation.'' \quad \textbf{B} (text): as in (b). \quad \textbf{C} (graph): no memory text; the KG enters only through the encoder. \quad \textbf{D} (text + graph): as in (b), plus the encoder. \quad \textbf{Per-node} (C, D): adds the line \texttt{\#\#\# Memory graph: <kg\_node>}$\times N$ after the memory block.
\end{minipage}}
\caption{The prompts, verbatim from the code. (a) MomaGraph-R1's prompt for each frame. (b) The answer prompt of the text model (arm B) for one test episode (S3, four rooms), with its memory rendered from our scene graphs; only the relation list is shortened. Red: errors found by comparing with the simulator's records. (c) How the prompt differs across arms. The navigation-action prompt is in App.~\ref{app:prompts}.}
\label{fig:prompts}
\end{figure*}

\textbf{Text.} The memory text lists the rooms in visit order and, under each room, at most 12 object types (those seen in the most frames) with their frame interval and count, followed by at most 24 relations.

\textbf{Graph encoder.} A relational graph network (R-GCN~\cite{schlichtkrull2018rgcn}) encodes the KG. Every node carries a learned name embedding plus state, function and timing features (the state and function slots stay empty with our prompt, which asks only for names), and every edge one of 12 relation types. The node embeddings are projected to the VLM's width.

\textbf{Standard graph route.} It combines two existing mechanisms (Fig.~\ref{fig:overview}). The \emph{pooled input} averages the node embeddings into one vector and adds it to every input position through a learned gate. The \emph{late cross-attention} lets the final hidden states attend to the node embeddings after the last layer, through a tanh gate that starts at zero~\cite{alayrac2022flamingo}. The first enters early but averages away individual nodes; the second keeps the nodes but enters after all reasoning layers. We train and test the two together as one configuration.

\textbf{Per-node route (ours).} The graph should enter early, so that every layer can use it, and per node, so that individual objects and their times survive. We replace both standard mechanisms with one input token per node (the graph encoder stays the same), as in LLaGA~\cite{chen2024llaga} and the object tokens of 3D language models~\cite{huang2024chatscene}. The answer prompt gains a line \texttt{\#\#\# Memory graph:} followed by $N$ placeholder tokens \tok. A hook on the token-embedding layer replaces the $k$-th placeholder with the $k$-th node embedding, passed through a two-layer projector and a LayerNorm and scaled to the size of ordinary token embeddings. Nodes keep the KG's order: rooms in visit order, then objects by first sighting. Unused placeholders get a learned padding vector, and attention masks and position codes need no change. The graph-only model has $N{=}64$ slots (median 32 nodes, at most 66); the text + graph model keeps walls and doors and has $N{=}80$ (median 36, at most 75). The new module's gradients were about 40 times larger than the LoRA adapter's, so we clip it separately (\appref{app:impl}).

\textbf{Fine-tuning.} Every arm is trained the same way (\appref{app:details}). LoRA adapters~\cite{hu2022lora} ($r{=}16$, $\alpha{=}32$) sit on the q, k, v and o projections; the graph encoder and its projections, where present, are trained from scratch with them. Each training sample is one step of a recorded route. At every step the model predicts the agent's next navigation action from the current frame and its scene graph (cross-entropy over the action set); at the final step it also answers the episode's questions (cross-entropy on the answer letter, answer steps oversampled $8\times$). Training runs 6,000 optimizer steps with batch 16 (AdamW, peak learning rate $10^{-4}$, cosine schedule) on the 5,000 training episodes. We evaluate the final checkpoint, never the best validation one, on the 800 test questions, scoring the answer by the most likely option letter. Differences are in accuracy points with 95\% intervals from a bootstrap over test episodes; * marks an interval that excludes zero, and two-seed results average per-question correctness over seeds first (\appref{app:stats}). Intervals are unadjusted for multiple comparisons; the comparisons behind the four findings are primary, and breakdowns by type, stratum and checkpoint are exploratory.

\section{Results}\label{sec:results}
\subsection{Does text memory help?}\label{sec:acc}

\begin{table}[t]
\centering
\caption{Test results by memory route: accuracy on the 800 test questions (\%) and the gain over no memory in accuracy points over all, route (291) and set (509) questions; * marks a paired 95\% interval that excludes zero (intervals in Table~\ref{tab:pairs}). Text, no memory and per-node models: two seeds; standard graph models: seed 0, compared with no memory seed 0. Shaded: our per-node route. Last row: text from perfect scene graphs, an upper bound.}
\label{tab:main}
\setlength{\tabcolsep}{2.3pt}
\footnotesize
\begin{tabular}{@{}l l c r r r@{}}
\toprule
 & & & \multicolumn{3}{c}{Gain over no memory}\\
\cmidrule(l){4-6}
Arm, memory & Route & Acc. & All & Route & Set\\
\midrule
A\enspace No memory & -- & 64.8 & \multicolumn{3}{c}{ref.}\\
B\enspace Text & -- & 75.9 & +11.06\sig & +30.07\sig & +0.20\\
\midrule
C\enspace Graph & standard & 65.9 & +1.62 & +3.44 & +0.59\\
\rowcolor{oursbg}C\enspace Graph & \textbf{per-node} & 67.1 & +2.31\sig & +8.25\sig & $-$1.08\\
\midrule
D\enspace Text + graph & standard & 77.0 & +12.75\sig & +27.49\sig & +4.32\sig\\
\rowcolor{oursbg}D\enspace Text + graph & \textbf{per-node} & 76.9 & +12.06\sig & +31.44\sig & +0.98\\
\midrule
{\color{dimgray}B\enspace Text, perfect} & -- & {\color{dimgray}83.4} & {\color{dimgray}+18.62\sig} & {\color{dimgray}+38.32\sig} & {\color{dimgray}+7.37\sig}\\
\bottomrule
\end{tabular}
\end{table}

\begin{figure}[t]
\centering
\includegraphics[width=\columnwidth]{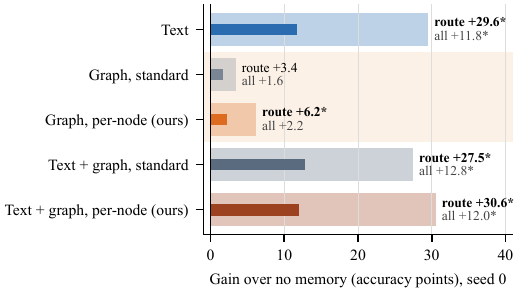}
\caption{Gain over no memory (seed 0), in accuracy points. Wide pale bar: route questions; narrow bar: all questions; * marks a paired 95\% interval that excludes zero. Of the graph-only models (shaded), only the per-node one clears zero, and only on route questions.}
\label{fig:headline}
\end{figure}

\begin{table*}[t]
\centering
\caption{Accuracy (\%) by question type on the 800 test questions. Top: reference models that are not trained on the benchmark; they answer from the question's evidence frames (the language-only model sees no image). Bottom: fine-tuned models that answer from memory with a blank image. The two regimes do not share an evidence protocol and are not one leaderboard. $S$: seeds averaged. Green (red): the paired 95\% interval of the difference to the same backbone's no-memory model excludes zero, with seeds matched. Bold: best in the column within each block. Shaded rows: our per-node route. Chance is question-weighted.}
\label{tab:big}
\footnotesize
\setlength{\tabcolsep}{3.1pt}
\renewcommand{\arraystretch}{1.04}
\begin{tabular}{@{}l l c ccc>{\columncolor{avgbg}}c ccccc>{\columncolor{avgbg}}c>{\columncolor{avgbg}}c@{}}
\toprule
 & & & \multicolumn{4}{c}{Route: when things were seen} & \multicolumn{6}{c}{Set: what was where} & \multicolumn{1}{c}{}\\
\cmidrule(lr){4-7}\cmidrule(lr){8-13}
Model & Memory, graph route & $S$ & Order & Pres. & Visited & Route & Persist & Room & Exist & Count & Global & Set & All\\
 & & & {\color{dimgray}\scriptsize 73} & {\color{dimgray}\scriptsize 102} & {\color{dimgray}\scriptsize 116} & {\color{dimgray}\scriptsize 291} & {\color{dimgray}\scriptsize 131} & {\color{dimgray}\scriptsize 80} & {\color{dimgray}\scriptsize 113} & {\color{dimgray}\scriptsize 65} & {\color{dimgray}\scriptsize 120} & {\color{dimgray}\scriptsize 509} & {\color{dimgray}\scriptsize 800}\\
\midrule
\multicolumn{3}{@{}l}{Chance (question-weighted)} & 50.0 & 50.0 & 28.9 & 41.6 & 30.0 & 28.2 & 25.1 & 28.6 & 28.6 & 28.1 & 33.0\\
\midrule
\multicolumn{14}{@{}l}{\textit{Language only, no image}}\\
\quad Qwen3-4B~\cite{qwen2025qwen3} & -- & -- & 39.7 & 51.0 & 30.2 & 39.9 & 38.2 & 70.0 & 61.1 & 26.2 & 32.5 & 45.4 & 43.4\\
\midrule
\multicolumn{14}{@{}l}{\textit{Zero-shot VLMs, evidence frames}}\\
\quad Qwen2.5-VL-32B~\cite{bai2025qwen25vl} & -- & -- & 78.1 & \textbf{52.0} & \textbf{72.4} & \textbf{66.7} & \textbf{71.0} & 66.2 & 69.9 & 49.2 & \textbf{31.7} & \textbf{58.0} & \textbf{61.1}\\
\quad Qwen3-VL-8B~\cite{qwen2025qwen3vl} & -- & -- & \textbf{80.8} & 49.0 & 60.3 & 61.5 & \textbf{71.0} & 65.0 & 67.3 & 43.1 & 30.0 & 56.0 & 58.0\\
\quad Qwen3-VL-4B~\cite{qwen2025qwen3vl} & -- & -- & 74.0 & 40.2 & 67.2 & 59.5 & 68.7 & 66.2 & \textbf{73.5} & \textbf{50.8} & 30.0 & \textbf{58.0} & 58.5\\
\quad Qwen2.5-VL-7B~\cite{bai2025qwen25vl} & -- & -- & 61.6 & 40.2 & 69.0 & 57.0 & 61.8 & \textbf{68.8} & 68.1 & 40.0 & 29.2 & 53.8 & 55.0\\
\quad LLaVA-1.6-7B~\cite{liu2024llavanext} & -- & -- & 63.0 & 49.0 & 23.3 & 42.3 & 67.2 & 38.8 & 68.1 & 41.5 & 23.3 & 49.3 & 46.8\\
\quad Idefics3-8B~\cite{laurencon2024idefics3} & -- & -- & 56.2 & 37.3 & 43.1 & 44.3 & 55.7 & 45.0 & 64.6 & 41.5 & 26.7 & 47.3 & 46.2\\
\midrule
\multicolumn{14}{@{}l}{\textit{Fine-tuned MomaGraph-R1, our scene graphs, answering from memory}}\\
\quad No memory & -- & 2 & 54.1 & 58.3 & 52.2 & 54.8 & 80.5 & 83.1 & 86.3 & 66.2 & 38.8 & 70.5 & 64.8\\
\quad Text & -- & 2 & \cellcolor{gainbg}80.8 & \cellcolor{gainbg}\textbf{94.1} & \cellcolor{gainbg}79.3 & \cellcolor{gainbg}84.9 & 80.9 & 83.1 & 87.6 & 67.7 & 37.1 & 70.7 & \cellcolor{gainbg}75.9\\
\quad Graph & standard & 1 & 63.0 & 58.8 & 56.9 & 59.1 & 80.2 & 81.2 & 87.6 & 61.5 & 38.3 & 69.7 & 65.9\\
\rowcolor{oursbg}\quad Graph & \textbf{per-node} & 2 & 54.8 & 60.8 & \cellcolor{gainbg}70.3 & \cellcolor{gainbg}63.1 & 80.9 & 81.2 & 88.1 & 65.4 & \cellcolor{lossbg}33.8 & 69.4 & \cellcolor{gainbg}67.1\\
\quad Text + graph & standard & 1 & \cellcolor{gainbg}78.1 & \cellcolor{gainbg}92.2 & \cellcolor{gainbg}78.4 & \cellcolor{gainbg}83.2 & \textbf{84.7} & 85.0 & 88.5 & \textbf{72.3} & \textbf{40.0} & \cellcolor{gainbg}\textbf{73.5} & \cellcolor{gainbg}\textbf{77.0}\\
\rowcolor{oursbg}\quad Text + graph & \textbf{per-node} & 2 & \cellcolor{gainbg}\textbf{84.2} & \cellcolor{gainbg}93.1 & \cellcolor{gainbg}\textbf{81.5} & \cellcolor{gainbg}\textbf{86.3} & 80.5 & \textbf{85.6} & \textbf{89.8} & 66.2 & 37.9 & 71.5 & \cellcolor{gainbg}76.9\\
\midrule
\multicolumn{14}{@{}l}{\textit{Fine-tuned MomaGraph-R1, perfect scene graphs (simulator records)}}\\
\quad Text & -- & 2 & \cellcolor{gainbg}93.8 & \cellcolor{gainbg}98.0 & \cellcolor{gainbg}88.4 & \cellcolor{gainbg}93.1 & 83.2 & 91.2 & \cellcolor{gainbg}95.1 & \cellcolor{gainbg}80.8 & 45.4 & \cellcolor{gainbg}77.9 & \cellcolor{gainbg}83.4\\
\bottomrule
\end{tabular}
\end{table*}

\textbf{Text sets the bar.} Text memory adds 11.06\sig\ points over no memory (two seeds; Table~\ref{tab:main}). The gain sits on route questions (+30.07\sig), with 27 to 36 points on each of the three route types (Fig.~\ref{fig:types}); on set questions text changes nothing (+0.20). Text from perfect scene graphs adds 18.62\sig, so better graph content still has room to help a text model.

\textbf{Overall accuracy hides where memory helps.} Fine-tuning teaches strong answer priors. The no-memory model answers 70.5\% of set questions, against 58.0\% for the best zero-shot VLM in Table~\ref{tab:big}, which sees the question's evidence frames, so any memory gain has to come on top of these priors. We therefore report route and set questions separately throughout.

\subsection{Why does the standard graph route not help?}\label{sec:std}
\textbf{The standard graph route stays at the no-memory level.} The graph-only model with the standard route gains 1.62 points, with an interval that includes zero, and 3.44 on route questions, also not significant. Its late-fusion gate ends at $\gate{=}0.062$ (\appref{app:logs}). The pooled input and the late cross-attention were trained only together, so this result cannot say which of the two blocks the graph; training each alone would separate them.

\textbf{Is the information lost in the encoder?} Two explanations fit this null result: the encoder discards the episode's facts, or they survive in the node embeddings and are lost on the way into the model. A logistic-regression probe tests one fact, first-sighting order. It predicts which of two object nodes was seen first from the encoder output of the standard graph model (8,000 node pairs from the 200 test episodes). It reaches AUC 0.807, against 0.495 (chance) for the raw node features with their time features hidden and 0.856 for an untrained encoder (\appref{app:probe}). The order can be read from the embeddings, so the loss happens after the encoder; that training leaves it slightly below a fresh encoder suggests the encoder receives little useful gradient through the standard route. Even so, the probe shows only that order can be read from the embeddings by a classifier trained for it; whether the VLM reads it is what the swap test in \secref{sec:swap} measures.

\subsection{Does one token per node make the graph usable?}\label{sec:pernode}
\textbf{The per-node route gets part of the graph in.} It lifts the graph-only model to a small but significant gain over no memory (+2.31\sig, two seeds) and to 8.25\sig\ points on route questions. On ``how many rooms did the robot visit'' it reaches 70.3\%, against 52.2\% without memory and 56.9\% with the standard route. The direct comparison with the standard route (seed 0) stays within noise, +1.25 over all questions and +3.95 on route questions (\appref{app:results}), so accuracy alone does not separate the two routes.

\textbf{With text present, the route does not change accuracy.} Both text + graph models gain about 12 points over no memory, the same as text alone (+1.00 for both against text, within noise).

\subsection{Does the model depend on its own graph?}\label{sec:swap}
\begin{figure}[t]
\centering
\includegraphics[width=\columnwidth]{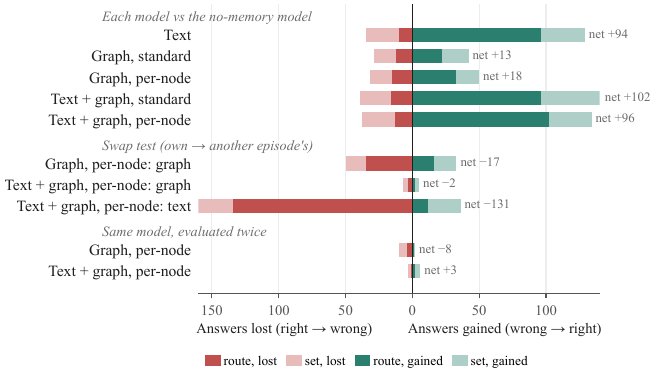}
\caption{Answers that change from right to wrong (left) or wrong to right (right), on route and set questions. Top: each model against no memory. Middle: the two per-node models with another episode's graph or text. Bottom: the same final checkpoint evaluated twice, the noise floor.}
\label{fig:flips}
\end{figure}

Accuracy cannot tell whether a graph route carries the episode's content or acts like a fixed prompt. The \emph{graph-swap test} scores each test episode with another episode's graph and leaves the question, the image and any memory text untouched. A \emph{text swap} does the same for the memory text. Both per-node models are tested at their final checkpoints, with own and swapped memory scored in one session (\appref{app:swap}); in that session they score 524 and 613, against 532 and 610 for the same seed-0 checkpoints in Table~\ref{tab:runs}, a difference within re-evaluation noise. The swap test uses seed 0 and one fixed derangement of the 200 test episodes; we have not measured how the cost varies across derangements (\appref{app:swap}). Fig.~\ref{fig:flips} counts the answers each change flips. As a noise floor, evaluating the same checkpoint twice changes 12 answers for the graph model and 9 for the text + graph model.

\textbf{The graph-only model depends on its graph.} With another episode's graph it loses 50 answers and gains 33, a net cost of 2.12 points overall and 6.53\sig\ on route questions. Its rooms-visited accuracy falls from 66.4\% to 47.4\% (Table~\ref{tab:swapboth}, \appref{app:swap}). Per-node tokens therefore deliver episode-specific information that this model uses.

\begin{figure}[t]
\centering
\includegraphics[width=\columnwidth]{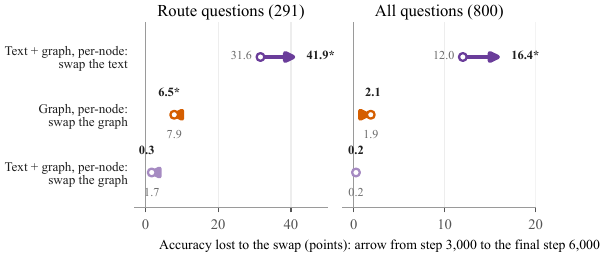}
\caption{Accuracy lost to each swap, per-node models: each arrow runs from the step-3,000 snapshot (open circle, gray value) to the final checkpoint (arrowhead, bold value). * marks a paired 95\% interval that excludes zero. The text-swap cost grows during training; the graph-swap cost of the text + graph model stays near zero.}
\label{fig:reliance}
\end{figure}

\textbf{The text + graph model shows no dependence on its graph.} The graph swap changes 12 answers (7 lost, 5 gained; 0.25 points), the same as the noise floor. The text swap changes 203 answers (167 lost, 134 of them on route questions; 36 gained). It costs 16.38\sig\ points and puts the model 4.00\sig\ points below no memory (482 vs 514), so the text route carries episode content rather than the mere presence of a memory field. This reliance on text grew during training. At step 3,000 the text swap cost 12.00 points, 4.38\sig\ less than at the end, while the graph swap cost 0.25 points at both checkpoints (Fig.~\ref{fig:reliance}). A null swap effect is a statement about this trained model under this intervention. The same per-node tokens carry usable content in the graph-only model; next to text, the model has learned not to need them.

\section{Failure Analysis}\label{sec:failures}
The gains above are averages. This section separates three limits on them: what perception puts into the memory, what the graph route lets the model read, and how both change with route length.

\begin{figure}[t]
\centering
\includegraphics[width=\columnwidth]{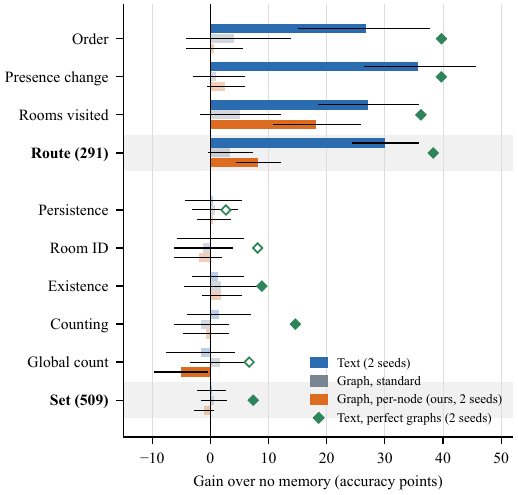}
\caption{Gain over no memory by question type. Bars: text and the per-node graph model (two seeds) and the standard graph model (seed 0); thin line: paired 95\% interval; pale bar: the interval includes zero. Diamonds: text from perfect scene graphs (two seeds; open: the interval includes zero). Shaded rows: route and set averages.}
\label{fig:types}
\end{figure}

\subsection{Perception limits the facts in memory}
\textbf{A typical memory.} Fig.~\ref{fig:prompts}b shows the memory of one test episode with its errors marked. The room changes are right, since they come from the door crossings, but the first room is misnamed, objects seen through a doorway are filed under the agent's room, a sofa is missed, and a few out-of-vocabulary names slip through. Route facts such as which object appeared first survive these errors; facts about what stands in which room do not.

\textbf{Set questions are limited by perception.} No model that reads our scene graphs gains significantly on set questions except the standard text + graph model (+4.32\sig, one seed; exploratory). Text from perfect graphs does: +7.37\sig\ on set questions, with +14.6\sig\ on counting and +8.9\sig\ on existence. Our graphs still miss half of the visible object types (recall 49.2) and file only 45.5\% of (type, room) pairs under the right room, so the facts set questions need are often absent from the memory. For a text model, better perception is the lever our data supports. Whether a graph-only model would gain as much from perfect graphs is untested, since no graph model was trained on them.

\subsection{The graph route limits what the model can read}
\textbf{The graph-only route carries room structure and little object order.} The per-node graph model's only significant gain on a single type is rooms visited (+18.1\sig; Fig.~\ref{fig:types}), and the graph swap removes it (\secref{sec:swap}). On order questions it is 0.7 points above no memory and on presence change 2.5 above, both within noise, while text gains 27 and 36 points there; on global counts it loses 5.0\sig\ points (exploratory; noisy gold answers, \appref{app:data}). The tokens deliver the sequence of room nodes, but the model does not yet read the first-sighting times of individual objects from them.

\textbf{Next to text, the graph adds nothing the model needs.} The text + graph model shows no dependence on its graph (\secref{sec:swap}). One explanation is redundancy: the answer loss is already met by the text, so a graph that repeats it gets no pressure to be used. We have not tested this. Training with the memory text randomly removed would test it; the switch is implemented but no model has been trained with it (\appref{app:impl}).

\subsection{Longer routes amplify both limits}
Route strata count door crossings: S1 one, S2 two, S3 three, S4 four or more. Each stratum has 50 test episodes and the split is post hoc, so we read the results below as a pattern rather than a test.

\begin{figure}[t]
\centering
\includegraphics[width=\columnwidth]{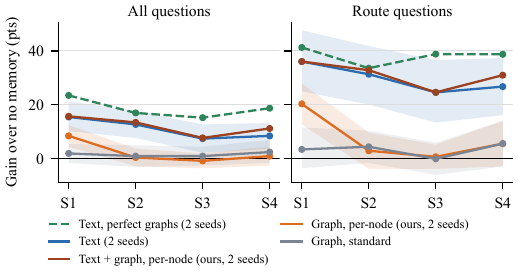}
\caption{Gain over no memory by number of door crossings (strata S1 to S4, 50 test episodes each), over all questions (left) and route questions (right). Bands: paired 95\% intervals for text and the two graph-only models. Text, perfect-graph text and per-node models: two seeds; standard graph: seed 0 against no memory seed 0.}
\label{fig:strata}
\end{figure}

\textbf{The graph-only gain is confined to short routes.} The per-node graph model gains 8.5\sig\ points in S1 (20.4\sig\ on route questions) and nothing in S2 or S3 (+0.5 and $-$0.8; Fig.~\ref{fig:strata}). Text gains in every stratum.

\textbf{Memory gains shrink on longer routes.} Text gains most in S1 (+15.5\sig) and about half as much in S3 and S4 (+7.5\sig, +8.5\sig). Text from perfect graphs keeps 15 to 24 points in every stratum, so on route questions the gap between our graphs and perfect ones grows from 5 points in S1 to 12 to 14 in S3 and S4. A likely cause is that errors in the predicted graphs accumulate along the route. The no-memory model also scores slightly higher on longer routes (63.0\% to 66.0\%), which narrows the room for gains.

\section{Discussion and Conclusion}\label{sec:conclusion}
\textbf{What the experiments show.} On one backbone and one benchmark, with our scene graphs: text memory raises accuracy, almost entirely on route questions; the standard graph route gives no clear gain; per-node tokens give a graph-only model a route-question gain that the graph swap removes; and with text present, the swap detects no dependence on the graph. Three practices follow directly. Report a graph-swap test next to accuracy and compare the answers it flips with re-evaluation noise; without it, a graph route that is never used looks like a modest success. Report route and set questions separately, since fine-tuned priors already answer most set questions. Measure the scene graphs against ground truth before blaming the route; on set questions our limit is recall.

\textbf{What the results suggest.} Giving each node its own input position may keep distinctions that a pooled summary averages away, though the direct comparison of the two routes is within noise. For set questions, better perception may matter more than a new route; our evidence for this comes from a text model on perfect graphs. A graph that repeats the text may go unused because the text already meets the loss. If so, a graph earns its place by carrying what the text lacks, such as 3D positions or the room an object actually stands in, or by training with the text sometimes absent.

\textbf{What remains open.} Other backbones, tasks and environments; a graph-only model on perfect graphs; memory-text dropout; and the swap test on other structured inputs.

\textbf{Limitations.} One simulator and one task family. One seed for the standard graph and text + graph models; two for the per-node models and the swap test on seed 0 only, with one derangement. All models use one backbone (MomaGraph-R1). The pooled input and the late cross-attention were tested together, not separately. Intervals resample episodes from 53 houses (33 give five each), so they may understate uncertainty for new houses, and they are unadjusted for multiple comparisons (\appref{app:stats}). Room segmentation uses the simulator's record of door crossings; a deployed robot would have to estimate them. The zero-shot rows of Table~\ref{tab:big} see evidence frames while fine-tuned models see a blank image, so the two are not one ranking. Gold answers follow a strip of six sampled frames, while some questions are worded about the whole route or house; global-count answers are noisy even for human annotators (\appref{app:data}).

\textbf{Conclusion.} In our benchmark, giving a VLM a structured scene graph did not by itself make the model use the graph's episode-specific content. Whether it did depended on how the graph entered the model and on whether the same facts were already in the text. A structured representation being available to a model is not the same as the model depending on it, and a gain in accuracy cannot tell the two apart. The graph-swap test can: it asks whether the model needs its own graph, and so separates reasoning over the map from a shortcut around it.

\bibliographystyle{IEEEtran}
\bibliography{main}

@article{hughes2022hydra,
  title={Hydra: A real-time spatial perception system for 3D scene graph construction and optimization},
  author={Hughes, Nathan and Chang, Yun and Carlone, Luca},
  journal={arXiv preprint arXiv:2201.13360},
  year={2022}
}

@inproceedings{gu2024conceptgraphs,
  title={Conceptgraphs: Open-vocabulary 3d scene graphs for perception and planning},
  author={Gu, Qiao and Kuwajerwala, Ali and Morin, Sacha and Jatavallabhula, Krishna Murthy and Sen, Bipasha and Agarwal, Aditya and Rivera, Corban and Paul, William and Ellis, Kirsty and Chellappa, Rama and others},
  booktitle={2024 IEEE International Conference on Robotics and Automation (ICRA)},
  pages={5021--5028},
  year={2024},
  organization={IEEE}
}

@inproceedings{werby2024hovsg,
  title={Hierarchical open-vocabulary 3d scene graphs for language-grounded robot navigation},
  author={Werby, Abdelrhman and Huang, Chenguang and B{\"u}chner, Martin and Valada, Abhinav and Burgard, Wolfram},
  booktitle={First Workshop on Vision-Language Models for Navigation and Manipulation at ICRA 2024},
  year={2024}
}

@article{rana2023sayplan,
  title={Sayplan: Grounding large language models using 3d scene graphs for scalable robot task planning},
  author={Rana, Krishan and Haviland, Jesse and Garg, Sourav and Abou-Chakra, Jad and Reid, Ian and Suenderhauf, Niko},
  journal={arXiv preprint arXiv:2307.06135},
  year={2023}
}

@article{yin2024sgnav,
  title={Sg-nav: Online 3d scene graph prompting for llm-based zero-shot object navigation},
  author={Yin, Hang and Xu, Xiuwei and Wu, Zhenyu and Zhou, Jie and Lu, Jiwen},
  journal={Advances in neural information processing systems},
  volume={37},
  pages={5285--5307},
  year={2024}
}

@inproceedings{fatemi2024talk,
  title={Talk like a graph: Encoding graphs for large language models},
  author={Fatemi, Bahare and Halcrow, Jonathan and Perozzi, Bryan},
  booktitle={International conference on learning representations},
  volume={2024},
  pages={43909--43934},
  year={2024}
}

@article{perozzi2024graphtoken,
  title={Let your graph do the talking: Encoding structured data for llms},
  author={Perozzi, Bryan and Fatemi, Bahare and Zelle, Dustin and Tsitsulin, Anton and Kazemi, Mehran and Al-Rfou, Rami and Halcrow, Jonathan},
  journal={arXiv preprint arXiv:2402.05862},
  year={2024}
}

@article{he2024gretriever,
  title={G-retriever: Retrieval-augmented generation for textual graph understanding and question answering},
  author={He, Xiaoxin and Tian, Yijun and Sun, Yifei and Chawla, Nitesh V and Laurent, Thomas and LeCun, Yann and Bresson, Xavier and Hooi, Bryan},
  journal={Advances in Neural Information Processing Systems},
  volume={37},
  pages={132876--132907},
  year={2024}
}

@article{chen2024llaga,
  title={Llaga: Large language and graph assistant},
  author={Chen, Runjin and Zhao, Tong and Jaiswal, Ajay and Shah, Neil and Wang, Zhangyang},
  journal={arXiv preprint arXiv:2402.08170},
  year={2024}
}

@inproceedings{tang2024graphgpt,
  title={Graphgpt: Graph instruction tuning for large language models},
  author={Tang, Jiabin and Yang, Yuhao and Wei, Wei and Shi, Lei and Su, Lixin and Cheng, Suqi and Yin, Dawei and Huang, Chao},
  booktitle={Proceedings of the 47th international ACM SIGIR conference on research and development in information retrieval},
  pages={491--500},
  year={2024}
}

@article{maggio2024clio,
  title={Clio: Real-time task-driven open-set 3d scene graphs},
  author={Maggio, Dominic and Chang, Yun and Hughes, Nathan and Trang, Matthew and Griffith, Dan and Dougherty, Carlyn and Cristofalo, Eric and Schmid, Lukas and Carlone, Luca},
  journal={IEEE Robotics and Automation Letters},
  volume={9},
  number={10},
  pages={8921--8928},
  year={2024},
  publisher={IEEE}
}

@inproceedings{huang2023vlmaps,
  title={Visual language maps for robot navigation},
  author={Huang, Chenguang and Mees, Oier and Zeng, Andy and Burgard, Wolfram},
  booktitle={2023 IEEE International Conference on Robotics and Automation (ICRA)},
  pages={10608--10615},
  year={2023},
  organization={IEEE}
}

@inproceedings{anwar2025remembr,
  title={Remembr: Building and reasoning over long-horizon spatio-temporal memory for robot navigation},
  author={Anwar, Abrar and Welsh, John and Biswas, Joydeep and Pouya, Soha and Chang, Yan},
  booktitle={2025 IEEE International Conference on Robotics and Automation (ICRA)},
  pages={2838--2845},
  year={2025},
  organization={IEEE}
}

@inproceedings{das2018eqa,
  title={Embodied question answering},
  author={Das, Abhishek and Datta, Samyak and Gkioxari, Georgia and Lee, Stefan and Parikh, Devi and Batra, Dhruv},
  booktitle={2018 IEEE/CVF Conference on Computer Vision and Pattern Recognition Workshops (CVPRW)},
  pages={2135--213509},
  year={2018},
  organization={IEEE}
}

@inproceedings{majumdar2024openeqa,
  title={Openeqa: Embodied question answering in the era of foundation models},
  author={Majumdar, Arjun and Ajay, Anurag and Zhang, Xiaohan and Putta, Pranav and Yenamandra, Sriram and Henaff, Mikael and Silwal, Sneha and Mcvay, Paul and Maksymets, Oleksandr and Arnaud, Sergio and others},
  booktitle={2024 IEEE/CVF Conference on Computer Vision and Pattern Recognition (CVPR)},
  pages={16488--16498},
  year={2024},
  organization={IEEE}
}

@article{ren2024explore,
  title={Explore until confident: Efficient exploration for embodied question answering},
  author={Ren, Allen Z and Clark, Jaden and Dixit, Anushri and Itkina, Masha and Majumdar, Anirudha and Sadigh, Dorsa},
  journal={arXiv preprint arXiv:2403.15941},
  year={2024}
}

@article{saxena2025grapheqa,
  title={Grapheqa: Using 3d semantic scene graphs for real-time embodied question answering},
  author={Saxena, Saumya and Buchanan, Blake and Paxton, Chris and Liu, Peiqi and Chen, Bingqing and Vaskevicius, Narunas and Palmieri, Luigi and Francis, Jonathan and Kroemer, Oliver},
  journal={arXiv preprint arXiv:2412.14480},
  year={2024}
}

@article{ali2025graphpad,
  title={GraphPad: Inference-Time 3D Scene Graph Updates for Embodied Question Answering},
  author={Ali, Muhammad Qasim and Nair, Saeejith and Wong, Alexander and Cui, Yuchen and Chen, Yuhao},
  journal={arXiv preprint arXiv:2506.01174},
  year={2025}
}

@inproceedings{yang2025mem,
  title={3d-mem: 3d scene memory for embodied exploration and reasoning},
  author={Yang, Yuncong and Yang, Han and Zhou, Jiachen and Chen, Peihao and Zhang, Hongxin and Du, Yilun and Gan, Chuang},
  booktitle={2025 IEEE/CVF Conference on Computer Vision and Pattern Recognition (CVPR)},
  pages={17294--17303},
  year={2025},
  organization={IEEE}
}

@article{hu2025mem,
  title={3dllm-mem: Long-term spatial-temporal memory for embodied 3d large language model},
  author={Hu, Wenbo and Hong, Yining and Wang, Yanjun and Gao, Leison and Wei, Zibu and Yao, Xingcheng and Peng, Nanyun and Bitton, Yonatan and Szpektor, Idan and Chang, Kai-Wei},
  journal={Advances in Neural Information Processing Systems},
  volume={38},
  pages={67856--67884},
  year={2026}
}

@inproceedings{yang2025thinking,
  title={Thinking in space: How multimodal large language models see, remember, and recall spaces},
  author={Yang, Jihan and Yang, Shusheng and Gupta, Anjali W and Han, Rilyn and Fei-Fei, Li and Xie, Saining},
  booktitle={2025 IEEE/CVF Conference on Computer Vision and Pattern Recognition (CVPR)},
  pages={10632--10643},
  year={2025},
  organization={IEEE}
}

@article{wang2023nlgraph,
  title={Can language models solve graph problems in natural language?},
  author={Wang, Heng and Feng, Shangbin and He, Tianxing and Tan, Zhaoxuan and Han, Xiaochuang and Tsvetkov, Yulia},
  journal={Advances in Neural Information Processing Systems},
  volume={36},
  pages={30840--30861},
  year={2023}
}

@inproceedings{tian2024gnp,
  title={Graph neural prompting with large language models},
  author={Tian, Yijun and Song, Huan and Wang, Zichen and Wang, Haozhu and Hu, Ziqing and Wang, Fang and Chawla, Nitesh V and Xu, Panpan},
  booktitle={Proceedings of the AAAI conference on artificial intelligence},
  volume={38},
  number={17},
  pages={19080--19088},
  year={2024}
}

@article{liu2023llava,
  title={Visual instruction tuning},
  author={Liu, Haotian and Li, Chunyuan and Wu, Qingyang and Lee, Yong Jae},
  journal={Advances in neural information processing systems},
  volume={36},
  pages={34892--34916},
  year={2023}
}

@article{hong20233dllm,
  title={3d-llm: Injecting the 3d world into large language models},
  author={Hong, Yining and Zhen, Haoyu and Chen, Peihao and Zheng, Shuhong and Du, Yilun and Chen, Zhenfang and Gan, Chuang},
  journal={Advances in Neural Information Processing Systems},
  volume={36},
  pages={20482--20494},
  year={2023}
}

@article{huang2024chatscene,
  title={Chat-scene: Bridging 3d scene and large language models with object identifiers},
  author={Huang, Haifeng and Chen, Yilun and Wang, Zehan and Huang, Rongjie and Xu, Runsen and Wang, Tai and Liu, Luping and Cheng, Xize and Zhao, Yang and Pang, Jiangmiao and others},
  journal={arXiv preprint arXiv:2312.08168},
  year={2023}
}

@article{zhang2022greaselm,
  title={Greaselm: Graph reasoning enhanced language models for question answering},
  author={Zhang, Xikun and Bosselut, Antoine and Yasunaga, Michihiro and Ren, Hongyu and Liang, Percy and Manning, Christopher D and Leskovec, Jure},
  journal={arXiv preprint arXiv:2201.08860},
  year={2022}
}

@article{alayrac2022flamingo,
  title={Flamingo: a visual language model for few-shot learning},
  author={Alayrac, Jean-Baptiste and Donahue, Jeff and Luc, Pauline and Miech, Antoine and Barr, Iain and Hasson, Yana and Lenc, Karel and Mensch, Arthur and Millican, Katherine and Reynolds, Malcolm and others},
  journal={Advances in neural information processing systems},
  volume={35},
  pages={23716--23736},
  year={2022}
}

@inproceedings{wang2020makes,
  title={What makes training multi-modal classification networks hard?},
  author={Wang, Weiyao and Tran, Du and Feiszli, Matt},
  booktitle={2020 IEEE/CVF Conference on Computer Vision and Pattern Recognition (CVPR)},
  pages={12692--12702},
  year={2020},
  organization={IEEE}
}

@inproceedings{wu2022greedy,
  title={Characterizing and overcoming the greedy nature of learning in multi-modal deep neural networks},
  author={Wu, Nan and Jastrzebski, Stanislaw and Cho, Kyunghyun and Geras, Krzysztof J},
  booktitle={International Conference on Machine Learning},
  pages={24043--24055},
  year={2022},
  organization={PMLR}
}

@inproceedings{agrawal2018vqacp,
  title={Don't just assume; look and answer: Overcoming priors for visual question answering},
  author={Agrawal, Aishwarya and Batra, Dhruv and Parikh, Devi and Kembhavi, Aniruddha},
  booktitle={2018 IEEE/CVF Conference on Computer Vision and Pattern Recognition},
  pages={4971--4980},
  year={2018},
  organization={IEEE}
}

@inproceedings{thomason2019shifting,
  title={Shifting the baseline: Single modality performance on visual navigation \& qa},
  author={Thomason, Jesse and Gordon, Daniel and Bisk, Yonatan},
  booktitle={Proceedings of the 2019 Conference of the North American Chapter of the Association for Computational Linguistics: Human Language Technologies, Volume 1 (Long and Short Papers)},
  pages={1977--1983},
  year={2019}
}

@article{geirhos2020shortcut,
  title={Shortcut learning in deep neural networks},
  author={Geirhos, Robert and Jacobsen, J{\"o}rn-Henrik and Michaelis, Claudio and Zemel, Richard and Brendel, Wieland and Bethge, Matthias and Wichmann, Felix A},
  journal={Nature Machine Intelligence},
  volume={2},
  number={11},
  pages={665--673},
  year={2020},
  publisher={Nature Publishing Group UK London}
}

@article{neverova2016moddrop,
  title={Moddrop: adaptive multi-modal gesture recognition},
  author={Neverova, Natalia and Wolf, Christian and Taylor, Graham and Nebout, Florian},
  journal={IEEE Transactions on Pattern Analysis and Machine Intelligence},
  volume={38},
  number={8},
  pages={1692--1706},
  year={2015},
  publisher={IEEE}
}

@article{deitke2022procthor,
  title={ProcTHOR: Large-Scale Embodied AI Using Procedural Generation},
  author={Deitke, Matt and VanderBilt, Eli and Herrasti, Alvaro and Weihs, Luca and Ehsani, Kiana and Salvador, Jordi and Han, Winson and Kolve, Eric and Kembhavi, Aniruddha and Mottaghi, Roozbeh},
  journal={Advances in neural information processing systems},
  volume={35},
  pages={5982--5994},
  year={2022}
}

@article{kolve2017ai2thor,
  title = {{AI2-THOR: An Interactive 3D Environment for Visual AI}},
  author = {Eric Kolve and Roozbeh Mottaghi and Winson Han and Eli VanderBilt and Luca Weihs and Alvaro Herrasti and Matt Deitke and Kiana Ehsani and Daniel Gordon and Yuke Zhu and Aniruddha Kembhavi and Abhinav Gupta and Ali Farhadi},
  journal = {arXiv preprint arXiv:1712.05474},
  year = {2017},
  url = {https://arxiv.org/abs/1712.05474}
}

@inproceedings{ju2026momagraph,
  title = {{MomaGraph: State-Aware Unified Scene Graphs with Vision-Language Models for Embodied Task Planning}},
  author = {Yuanchen Ju and Yongyuan Liang and Yen-Jen Wang and Nandiraju Gireesh and Yuanliang Ju and Seungjae Lee and Qiao Gu and Elvis Hsieh and Furong Huang and Koushil Sreenath},
  booktitle = {International Conference on Learning Representations (ICLR)},
  year = {2026},
  url = {https://arxiv.org/abs/2512.16909}
}

@inproceedings{schlichtkrull2018rgcn,
  title = {{Modeling Relational Data with Graph Convolutional Networks}},
  author = {Michael Schlichtkrull and Thomas N. Kipf and Peter Bloem and Rianne van den Berg and Ivan Titov and Max Welling},
  booktitle = {The Semantic Web: 15th International Conference, ESWC 2018},
  series = {Lecture Notes in Computer Science},
  volume = {10843},
  pages = {593--607},
  publisher = {Springer},
  year = {2018},
  doi = {10.1007/978-3-319-93417-4_38},
  url = {https://arxiv.org/abs/1703.06103}
}

@article{bai2025qwen25vl,
  title = {{Qwen2.5-VL Technical Report}},
  author = {Shuai Bai and Keqin Chen and Xuejing Liu and Jialin Wang and Wenbin Ge and Sibo Song and Kai Dang and Peng Wang and Shijie Wang and Jun Tang and Humen Zhong and Yuanzhi Zhu and others},
  journal = {arXiv preprint arXiv:2502.13923},
  year = {2025},
  url = {https://arxiv.org/abs/2502.13923}
}

@inproceedings{hu2022lora,
  title = {{LoRA: Low-Rank Adaptation of Large Language Models}},
  author = {Edward J. Hu and Yelong Shen and Phillip Wallis and Zeyuan Allen-Zhu and Yuanzhi Li and Shean Wang and Lu Wang and Weizhu Chen},
  booktitle = {International Conference on Learning Representations (ICLR)},
  year = {2022},
  url = {https://arxiv.org/abs/2106.09685}
}

@article{qwen2025qwen3,
  title = {{Qwen3 Technical Report}},
  author = {{Qwen Team}},
  journal = {arXiv preprint arXiv:2505.09388},
  year = {2025},
  url = {https://arxiv.org/abs/2505.09388}
}

@article{qwen2025qwen3vl,
  title = {{Qwen3-VL Technical Report}},
  author = {{Qwen Team}},
  journal = {arXiv preprint arXiv:2511.21631},
  year = {2025},
  url = {https://arxiv.org/abs/2511.21631}
}

@article{laurencon2024idefics3,
  title = {{Building and Better Understanding Vision-Language Models: Insights and Future Directions}},
  author = {Hugo Lauren{\c{c}}on and Andr{\'e}s Marafioti and Victor Sanh and L{\'e}o Tronchon},
  journal = {arXiv preprint arXiv:2408.12637},
  year = {2024},
  url = {https://arxiv.org/abs/2408.12637}
}

@misc{liu2024llavanext,
  title = {{LLaVA-NeXT: Improved Reasoning, OCR, and World Knowledge}},
  author = {Haotian Liu and Chunyuan Li and Yuheng Li and Bo Li and Yuanhan Zhang and Sheng Shen and Yong Jae Lee},
  year = {2024},
  month = jan,
  howpublished = {\url{https://llava-vl.github.io/blog/2024-01-30-llava-next/}}
}

\clearpage
\appendices

\section{Benchmark and Dataset Details}\label{app:data}
This section adds statistics and examples to the benchmark description in \secref{sec:bench} and Table~\ref{tab:qtypes}. Test statistics are computed from the test episode records.

\begin{figure}[t]
\centering
\includegraphics[width=0.32\columnwidth]{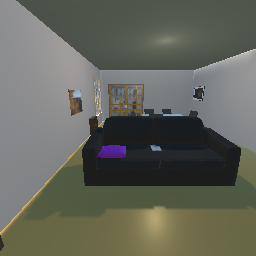}\hfill
\includegraphics[width=0.32\columnwidth]{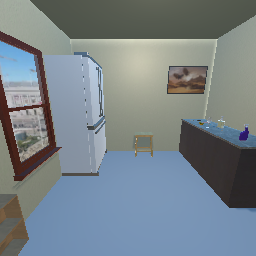}\hfill
\includegraphics[width=0.32\columnwidth]{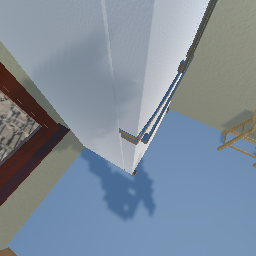}
\caption{Stored $256{\times}256$ frames from one episode outside the test split (31 frames, living room\,$\rightarrow$\,kitchen). Left: an early view in the living room. Middle: the view at the room transition, with the kitchen ahead. Right: a late view near the end of the route.}
\label{fig:s-example}
\end{figure}

\textbf{Episodes.} The route starts out of view of the target object and leads to it. The actions are MoveAhead by 0.25, 0.5 or 1.0\,m, RotateLeft/RotateRight by 15, 30, 45 or 90$^\circ$, LookUp/LookDown by 15 or 30$^\circ$ within a pitch range of $-30$ to 60$^\circ$, and Done. Each step also stores the agent pose. All 200 test routes reach their target with these discrete actions alone. Fig.~\ref{fig:s-example} shows views from one episode outside the test split.

\textbf{Visible-object record.} Each step also stores the simulator's record of visible objects: type, instance ID and room of each one. An object enters a frame's record when AI2-THOR flags it visible within 6\,m and its visible region covers at least 0.05\% of the image (about 33 pixels), so partly occluded and cut-off objects count. The room in each entry is the room the object stands in, which can differ from the agent's room when the object is seen through a doorway. Test frames contain 6.1 visible objects on average (median 5). A test episode shows 46.5 distinct object instances of 26.5 types on average.

\begin{figure}[t]
\centering
\includegraphics[width=\columnwidth]{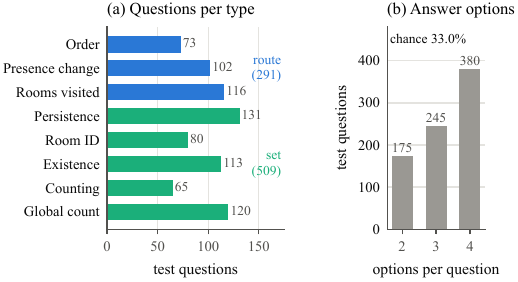}
\caption{Test-split composition. (a) Questions per type. (b) Options per question; the mix of two, three and four options puts uniform-guess accuracy at 33.0\%, above the 25\% of four options.}
\label{fig:s-dataset}
\end{figure}

\textbf{Questions and answers.} No language model writes questions or answers. Templates write four questions per episode from the records of six sampled frames, the \emph{question strip}, and every gold answer is computed from the strip's records and room labels (Table~\ref{tab:qtypes} in the main text). Answers therefore follow the strip even when a question's wording refers to the whole run or house. Each question cites one to four strip frames, and its options are shuffled. The test split has 175 two-option, 245 three-option and 380 four-option questions, so guessing uniformly scores 33.0\% (Fig.~\ref{fig:s-dataset}). The memory may use the spawn frames and all frames before the question is asked. \textbf{Global counts are noisy.} In a check with four annotators who saw whole episodes, their majority vote on global-count questions was at chance (31.2\% against 28.0\%), and agreement between annotators was low ($\kappa$ 0.11 to 0.18). Global-count scores, which stay between 33\% and 46\% for every fine-tuned model, should be read with that in mind.

\textbf{Splits and route lengths.} The houses come from ProcTHOR-10k's own training, validation and test partitions. A test house contributes one to five episodes (33 of the 53 contribute five). Each route stratum holds 50 test episodes (Table~\ref{tab:s-strata}). Room labels name room types, so the 62 test routes that pass two rooms of the same type, including all of S4, contain two rooms with one label. Across test frames, 44\% are in bedrooms and 17 to 21\% in each of the other three room types; the target room is a bathroom in 89 of the 200 test episodes. The splits hold 206,899 training, 21,982 validation and 8,335 test frames. The memory never includes the final frame of an episode, at which the questions are asked, so 8,135 test frames feed the memory; scene-graph scores in \secref{sec:sg} and \appref{app:sg} use these.

\begin{table}[t]
\centering
\caption{Test split by route stratum. Doors: door crossings on the route. Rooms: distinct rooms on the route. Frames: mean (range). Path: walked length (m). Objects: distinct object instances seen per episode. Repeat: episodes passing two rooms of the same type.}
\label{tab:s-strata}
\footnotesize
\setlength{\tabcolsep}{3.2pt}
\begin{tabular}{@{}lrccrrrr@{}}
\toprule
Stratum & Eps. & Doors & Rooms & Frames & Path & Objects & Repeat\\
\midrule
S1 & 50 & 1 & 2 & 20.4 (10--40) & 6.6 & 28.0 & 0\\
S2 & 50 & 2 & 2--3 & 33.6 (18--58) & 11.2 & 43.5 & 2\\
S3 & 50 & 3 & 4 & 46.4 (26--87) & 16.1 & 46.2 & 10\\
S4 & 50 & 4--6 & 5--7 & 66.2 (34--116) & 22.9 & 68.2 & 50\\
\bottomrule
\end{tabular}
\end{table}

\FloatBarrier
\section{Scene-Graph Construction}\label{app:sg}
This section gives every step that turns the agent's frames into the scene graphs of \secref{sec:sg} and scores the result on the test split (200 episodes, the 8,135 memory frames, scored by object type against the simulator's visible-object records). Fitted choices use only the training and validation splits.

\subsection{Per-frame prediction}
MomaGraph-R1~\cite{ju2026momagraph} is Qwen2.5-VL-7B trained with reinforcement learning to output task-oriented scene graphs. Given one frame and a prompt, it returns JSON with a scene type, object nodes and edges between object pairs (spatial relations from left of, right of, in front of, behind, higher than, lower than, close, far and touching). The room label of the frame is read from the scene type. Our closed-vocabulary prompt (Fig.~\ref{fig:prompts}a, App.~\ref{app:prompts}) asks only for object names, so the state and function fields of the graph schema stay empty. Every split is processed once and cached, so training and evaluation read identical graphs.

\subsection{Pipeline}
\textbf{Step 1: resolution.} Every frame is rendered in AI2-THOR at $512{\times}512$ from the agent pose and camera settings stored with the episode. Rendering at $256{\times}256$ with the same settings reproduces the stored frames (mean absolute pixel difference 0.39 of 255), so the viewpoints match the recorded episode. Rendering the 8,335 test frames takes 7.4 minutes on one GPU (Table~\ref{tab:hardware}).

\textbf{Step 2: closed-vocabulary prompt.} The prompt lists the benchmark's 104 object types plus Shelf and ShelvingUnit (frequent in the houses but missing from the question vocabulary), asks for small objects, and tells the model to leave out walls, doors and floors. On test, 7.5\% of the predicted mentions are still architecture and 3\% of the names fall outside the vocabulary; both are dropped.

\textbf{Step 3: per-type rules.} Because false positives repeat by type, each predicted type gets one rule: keep it, drop it, or rename it to another type of its family. Rules are added greedily, each time the one that most raises object-type $F_{0.5}$ (precision weighted above recall) on the validation split, until no rule helps. The fit returns 28 rules: three renames (Desk, Table and CoffeeTable to DiningTable) and 25 drops (Apple, Bathtub, Bottle, Cabinet, Cup, Desk\-Lamp, Fork, Knife, Lamp, Laptop, Mirror, Ottoman, Pen, Plate, RemoteControl, SideTable, SinkBasin, SoapBar, SoapBottle, Spoon, Stool, TableTopDecor, TissueBox, ToiletPaper, Vase). The final graphs reach object-type precision 65.7 at recall 49.2, and 29\% of their false positives are types absent from the whole episode.

\textbf{Step 4: rooms at door crossings.} Instead of a room guess per frame, the frames between two consecutive door crossings of the agent form one room segment. The crossings come from the simulator's record of the route; a deployed robot would have to estimate them from odometry and door detection. A logistic-regression classifier names each segment from two inputs: the calibrated object types seen in it and the counts of the VLM's per-frame room guesses. It is trained on the segments of the training split, with its regularisation chosen on validation. When a room type repeats along a route, later segments are numbered (``Bedroom 2'') so that the KG keeps them as separate rooms. The room label is right in 71.8\% of frames, the number of visited rooms in 199 of 200 episodes, and the edit distance between the remembered and the true room sequence is 1.30.

\subsection{Perfect scene graphs}
The perfect graphs are built from the same frames but from the episode's simulator records: the visible objects of each frame with their types, stable instance identities, states and rooms; the true room of each frame; and the spatial relations recorded for each frame. They need no prediction and are correct by construction. Only text models are trained on them (two seeds).

\subsection{From scene graphs to the episode KG}
For each question the KG is built from the frames the model may use: the spawn frames and every frame before the question. Rooms become nodes in visit order. An object type seen in a room becomes one object node keyed by type and room, carrying the first and last frame index in which it was seen and the largest number of copies seen in one frame. Relations between object types seen in the same frame become edges. The memory text renders this KG as rooms in visit order and, per room, at most 12 object types ranked by the number of frames in which they were seen (ties broken by the earliest sighting), each with its frame interval and, when above one, its count, followed by at most 24 distinct relations; walls, doors and floors are left out. The graph encoder reads the same KG as a graph: every node carries a 64-dimensional learned name embedding and 21 state, function and timing features, and every edge one of 12 relation types. Walls and doors are dropped for graph-only models and kept for text + graph models (Table~\ref{tab:train}). The per-node route orders nodes as rooms in visit order, then objects by first sighting.

\section{Prompts}\label{app:prompts}
Fig.~\ref{fig:prompts} in the main text gives the per-frame scene-graph prompt and the answer prompt. The 106 object types listed in the scene-graph prompt are the benchmark's 104 non-architectural types plus Shelf and ShelvingUnit: AlarmClock, Apple, ArmChair, BaseballBat, BasketBall, Bathtub, Bed, Book, Boots, Bottle, Bowl, Box, Bread, ButterKnife, CD, Cabinet, Candle, Cart, CellPhone, Chair, Cloth, ClothesDryer, CoffeeMachine, CoffeeTable, CounterTop, CreditCard, Cup, Desk, DeskLamp, Desktop, DiningTable, DishSponge, DogBed, Drawer, Dresser, Dumbbell, Egg, Faucet, FloorLamp, Footstool, Fork, Fridge, GarbageBag, GarbageCan, HousePlant, Kettle, KeyChain, Knife, Ladle, Lamp, Laptop, LaundryHamper, Lettuce, LightSwitch, Microwave, Mirror, Mug, Newspaper, Ottoman, Oven, Painting, Pan, PaperTowelRoll, Pen, Pencil, PepperShaker, Pillow, Plate, Plunger, Pot, Potato, RemoteControl, RoomDecor, Safe, SaltShaker, Shelf, ShelvingUnit, ShowerCurtain, SideTable, Sink, SinkBasin, SoapBar, SoapBottle, Sofa, Spatula, Spoon, SprayBottle, Statue, Stool, StoveBurner, TVStand, Table, TableTopDecor, TeddyBear, Television, TennisRacket, TissueBox, Toaster, Toilet, ToiletPaper, Tomato, VacuumCleaner, Vase, WashingMachine, Watch, WineBottle.

\textbf{Navigation-action prompt.} At every route step, all arms receive the current frame and this prompt, where the scene graph is the frame's own predicted graph and the action token's hidden state feeds the action classifier:

{\ttfamily\footnotesize\raggedright
\#\#\# Task: Predict the next navigation action.\\
\#\#\# Question: \{question\}\\
\#\#\# Choices: \{options\}\\
\#\#\# Candidate Actions: (0) \{action$_0$\} (1) \{action$_1$\} \dots\\
\#\#\# Predicted Scene Graph:\\
\{ROOM / OBJECTS / RELATIONS of the current frame\}\\
\#\#\# Guidance:\\
Use the scene graph and the question to choose the best next action.\\
Choose Done only when sufficient information has been gathered to answer the question.\\
\#\#\# Action Token: <CLS>\par}

\section{Experimental Details}\label{app:details}

\subsection{Model and training}
Table~\ref{tab:train} lists the training and evaluation settings. They are identical for every run; only the memory and the graph route change (and the gate warm-up, as listed).
\begin{table}[t]
\centering
\caption{Training and evaluation settings, shared by every run.}
\label{tab:train}
\footnotesize
\begin{tabular}{@{}>{\raggedright\arraybackslash}p{0.25\columnwidth}>{\raggedright\arraybackslash}p{0.69\columnwidth}@{}}
\toprule
Backbone & MomaGraph-R1 (Qwen2.5-VL-7B), bf16\\
Adapter & LoRA $r{=}16$, $\alpha{=}32$, dropout 0.05, on q, k, v, o projections\\
Optimiser & AdamW, peak lr $10^{-4}$, cosine schedule, 5\% warm-up, floor $0.05\times$, weight decay 0.01 (not on norms, biases, gates), gradient clip 1.0\\
Batch & $1\times16$ gradient accumulation; 6,000 optimizer steps\\
Losses & navigation-action cross-entropy on every step; answer-letter cross-entropy on the final step; answer frames oversampled $8\times$; answer weight 1.0\\
Answer image & blank (all episodic information comes from memory)\\
Graph encoder & 2-layer R-GCN over relation types; node features = 64-d name embedding + state and function features; hidden 128, output 256; projected to the VLM width (3,584) with Linear + LayerNorm\\
Gate warm-up & additive scaffold on the late-fusion gate: $0\to0.2$ over $[0,w]$, back to 0 over $[w,3w]$; $w{=}2{,}000$ for graph-only models, 400 for text + graph models\\
Architecture nodes & walls and doors dropped for graph-only models, kept for text + graph models\\
Evaluation & final checkpoint (step 6,000, never the best validation checkpoint); constrained scoring over option letters; 800 test questions\\
\bottomrule
\end{tabular}
\end{table}

\subsection{Linear probe}\label{app:probe}
Pairs of object nodes from the same test episode are labelled by whether node A was first seen before node B: 8,000 pairs over the 200 test episodes, at most 40 per episode. A logistic regression on the concatenated pair embedding is scored by AUC. Three inputs are compared: the raw node features with the explicit time block masked (otherwise the label is one of the features), a freshly initialised encoder, and the encoder of a trained graph model (arm C, seed 0, step 6,000, standard and per-node route). The random encoder is not seeded, so its AUC varies slightly between runs (0.834 to 0.856). The script is \texttt{code/probe\_kg.py}.

\begin{table}[t]
\centering
\caption{Linear-probe AUC for first-sighting order (8,000 test pairs).}
\label{tab:probe}
\footnotesize
\begin{tabular}{@{}lccc@{}}
\toprule
Encoder taken from & Raw & Random & Trained\\
\midrule
Graph, standard route (527) & 0.495 & 0.856 & 0.807\\
Graph, per-node route (532) & 0.495 & 0.834 & 0.834\\
\bottomrule
\end{tabular}
\end{table}

\subsection{Statistics}\label{app:stats}
All differences are paired over the same 800 test questions, with 95\% intervals from an episode bootstrap: 20,000 resamples of the 200 test episodes, keeping each episode's questions together, RNG seed 20260920. Where two seeds exist, per-question correctness is averaged over seeds before resampling. A difference is called significant when its interval excludes zero. To test whether a swap cost changed between two checkpoints, the same bootstrap is applied to the per-question difference of the two costs. Accuracies are given to one decimal and paired differences to two.

\textbf{Primary and exploratory comparisons.} The conclusions rest on six comparisons, those behind the four findings of \secref{sec:intro}: text against no memory; each graph-only model against no memory; and, in the swap test, the graph-only model's graph-swap cost and the text + graph model's graph-swap and text-swap costs. All other differences in Table~\ref{tab:pairs}, Table~\ref{tab:big} and Figs.~\ref{fig:types} and~\ref{fig:strata} (by question type, by stratum, between checkpoints) are exploratory. No interval is adjusted for multiple comparisons, so among the many exploratory intervals some will exclude zero by chance; we read a single significant subgroup, such as the standard text + graph model's set-question gain, as a lead for further runs. The text gain and the text-swap cost have intervals far from zero (lower bounds +8.50 and +13.25 points). Three primary results lie closer to zero and are the ones a stricter procedure could overturn: the per-node graph-only gain over all questions (+2.31\ci{+0.56}{+4.06}), its route-question gain (+8.25\ci{+4.47}{+12.03}) and its graph-swap cost on route questions (+6.53\ci{+1.79}{+11.26}).

\textbf{Episodes from the same house.} The bootstrap resamples the 200 test episodes, which come from 53 houses; 33 houses contribute five episodes each. Episodes in one house share rooms and objects, so their outcomes may be correlated. If they are, the episode-level intervals describe uncertainty over episodes in these houses and are too narrow for a claim about new houses. A house-level cluster bootstrap, resampling the 53 houses with all their episodes, would answer this; we have not run it, and the results near zero listed above are the ones it could change.

\FloatBarrier
\section{Graph-Swap Test}\label{app:swap}
\begin{table*}[t]
\centering
\caption{Graph-swap and text-swap test by question type (\%), at the step-3,000 snapshots and at the final step-6,000 checkpoints. Swapped conditions score each episode with another episode's graph, text, or both.}
\label{tab:swapboth}
\footnotesize
\setlength{\tabcolsep}{4pt}
\begin{tabular}{@{}l r cc cccc c cc cccc@{}}
\toprule
& & \multicolumn{6}{c}{Step 3,000} & & \multicolumn{6}{c}{Step 6,000 (final)}\\
\cmidrule(lr){3-8}\cmidrule(lr){10-15}
& & \multicolumn{2}{c}{Graph, per-node} & \multicolumn{4}{c}{Text + graph, per-node} & & \multicolumn{2}{c}{Graph, per-node} & \multicolumn{4}{c}{Text + graph, per-node}\\
\cmidrule(lr){3-4}\cmidrule(lr){5-8}\cmidrule(lr){10-11}\cmidrule(lr){12-15}
Type & $n$ & own & graph sw. & own & graph sw. & text sw. & both sw. & & own & graph sw. & own & graph sw. & text sw. & both sw.\\
\midrule
Order & 73 & 52.1 & 49.3 & 76.7 & 76.7 & 47.9 & 47.9 & & 54.8 & 56.2 & 84.9 & 83.6 & 43.8 & 43.8\\
Presence change & 102 & 52.9 & 52.0 & 87.3 & 86.3 & 46.1 & 49.0 & & 58.8 & 60.8 & 93.1 & 94.1 & 42.2 & 42.2\\
Rooms visited & 116 & 68.1 & 50.9 & 77.6 & 74.1 & 52.6 & 52.6 & & 66.4 & 47.4 & 81.9 & 81.0 & 47.4 & 48.3\\
\midrule
Persistence & 131 & 81.7 & 79.4 & 81.7 & 79.4 & 80.9 & 79.4 & & 78.6 & 80.9 & 80.9 & 80.2 & 79.4 & 80.2\\
Room ID & 80 & 77.5 & 78.8 & 76.2 & 77.5 & 75.0 & 75.0 & & 81.2 & 80.0 & 83.8 & 85.0 & 81.2 & 81.2\\
Existence & 113 & 85.0 & 85.0 & 88.5 & 88.5 & 86.7 & 87.6 & & 86.7 & 88.5 & 88.5 & 89.4 & 84.1 & 84.1\\
Counting & 65 & 67.7 & 66.2 & 67.7 & 70.8 & 70.8 & 70.8 & & 64.6 & 66.2 & 66.2 & 66.2 & 66.2 & 66.2\\
Global count & 120 & 30.0 & 39.2 & 38.3 & 40.8 & 36.7 & 39.2 & & 32.5 & 30.0 & 37.5 & 35.8 & 37.5 & 36.7\\
\midrule
All (correct of 800) & 800 & 516 & 501 & 593 & 591 & 497 & 502 & & 524 & 507 & 613 & 611 & 482 & 483\\
\bottomrule
\end{tabular}
\end{table*}

\subsection{Protocol}
A copy of the evaluation script adds one switch. Each test episode's graph is replaced by the graph of another test episode, chosen by a fixed permutation in which no episode keeps its own (RNG 20260920); the text swap uses the same mapping. All swap results therefore come from this one derangement. Each swap cost averages over 200 donor pairings, but a different derangement would pair episodes differently, and we have not measured the spread of the cost over several derangements. The derangement ignores houses, so a donor can come from the same house; such a donor shares rooms with the episode, which can only make the swap milder. Question, options and the blank image are unchanged. For the text + graph model we also run the text swap alone and both swaps together. Both per-node models were evaluated at their final step-6,000 checkpoints and, earlier, at step-3,000 snapshots copied out of the runs so that training could not overwrite them, on all 800 test questions (Fig.~\ref{fig:flips}, Table~\ref{tab:swapboth}). The original evaluation script is untouched; the change is a short diff (\texttt{code/eqa\_diag\_graphswap.diff}).

\subsection{Dependence on graph and text during training}\label{app:swapfinal}
Fig.~\ref{fig:reliance} in the main text compares the swap costs at the two checkpoints. The text + graph model's text-swap cost grew by 4.38\ci{+1.62}{+7.12}\sig\ points overall and by 10.31\ci{+5.28}{+15.47}\sig\ on route questions (from 31.62 to 41.92). Its graph-swap cost stayed below 2 points, with intervals that include zero, at both checkpoints. The graph-only model's graph-swap cost on route questions did not change significantly (7.90 and 6.53; difference $-$1.37\ci{$-$4.84}{+2.04}). Training made the text + graph model lean further on its text, and the swap detected no use of its graph at either checkpoint.

Table~\ref{tab:swapboth} gives the swap results by type. The graph-only model's dependence on its own graph is concentrated in rooms visited at both checkpoints. At step 3,000 the swapped graph scores higher on global count (39.2 vs 30.0), a sign that the graph-only route still pulls set answers away from the question prior. The text + graph model's route answers fall below the no-memory level only when the text is swapped, alone or with the graph (route questions $-$11.00\ci{$-$17.47}{$-$4.59}\sig\ against no memory).

\textbf{Evaluation noise.} Re-evaluating the same final checkpoint with identical settings changed 12 of 800 answers for the graph-only model (10 lost, 2 gained; 524 vs the 532 in Table~\ref{tab:main}) and 9 for the text + graph model (3 lost, 6 gained; 613 vs 610), most likely from non-deterministic bf16 GPU arithmetic on close answers. Differences below about one point between separate evaluation runs should not be interpreted. The own and swapped conditions come from one session, so this shift between sessions does not enter the swap comparisons; variation of the same size within a session may, which is why Fig.~\ref{fig:flips} shows it as a noise floor.

\FloatBarrier
\section{Full Results}\label{app:results}
All values are correct answers of 800 test questions. Predictions for every run are in \texttt{log\_corl/test\_predictions/}. Table~\ref{tab:pairs} lists every paired difference used in the paper, split into route and set questions.

\subsection{Runs on our scene graphs}
\begin{table}[t]
\centering
\caption{Correct answers of 800, our scene graphs. No memory reads no scene graph.}
\label{tab:ours}
\setlength{\tabcolsep}{3.5pt}
\footnotesize
\begin{tabular}{@{}l l c c@{}}
\toprule
Memory & Graph route & Seed 0 & Seed 1\\
\midrule
No memory & -- & 514 (64.2) & 523 (65.4)\\
Text & -- & 608 (76.0) & 606 (75.8)\\
Graph & standard & 527 (65.9) & --\\
Text + graph & standard & 616 (77.0) & --\\
Graph & per-node & 532 (66.5) & 542 (67.8)\\
Text + graph & per-node & 610 (76.2) & 620 (77.5)\\
\bottomrule
\multicolumn{4}{@{}l}{\scriptsize Accuracy (\%) in brackets.}
\end{tabular}
\end{table}

Table~\ref{tab:ours} gives every run on our scene graphs. Table~\ref{tab:big} in the main text gives the per-type accuracies. Text memory carries the route types for every model that reads it; the graph-only models gain only on rooms visited, and only with the per-node route. The only per-node difference to the standard route that clears zero is a set-question loss of the per-node text + graph model ($-$2.95\sig; Table~\ref{tab:pairs}).

\subsection{Perfect scene graphs (text model)}
Retrained on perfect scene graphs, the text model answers 663 and 672 questions (validation accuracy 86.3\% at step 6,000 for both seeds), 7.56\sig\ points above text from our graphs over all questions and 7.17\sig\ on set questions (Table~\ref{tab:pairs}). Per type, it gains significantly over no memory on all three route types and on existence and counting (Table~\ref{tab:big}); persistence does not move (83.2\% against 80.5\%).

\begin{table*}[t]
\centering
\caption{Every paired difference used in the paper, in accuracy points with 95\% episode-bootstrap intervals (* marks an interval that excludes zero). Two-seed rows average per-question correctness over seeds first.}
\label{tab:pairs}
\footnotesize
\setlength{\tabcolsep}{4pt}
\begin{tabular}{@{}l@{\hspace{14pt}}r@{\hspace{5pt}}l@{\hspace{14pt}}r@{\hspace{5pt}}l@{\hspace{14pt}}r@{\hspace{5pt}}l@{}}
\toprule
Comparison & \multicolumn{2}{c}{All questions (800)} & \multicolumn{2}{c}{Route questions (291)} & \multicolumn{2}{c}{Set questions (509)}\\
\midrule
\multicolumn{7}{@{}l}{\textit{Each model $-$ no memory (two seeds where both have two; otherwise seed 0)}}\\
\quad Text $-$ no memory, seed 0 & +11.75 & \llap{\sig\hspace{1.5pt}}\cit{+8.88}{+14.75} & +29.55 & \llap{\sig\hspace{1.5pt}}\cit{+23.57}{+35.49} & +1.57 & \cit{$-$1.31}{+4.44}\\
\quad Text $-$ no memory, two seeds & +11.06 & \llap{\sig\hspace{1.5pt}}\cit{+8.50}{+13.62} & +30.07 & \llap{\sig\hspace{1.5pt}}\cit{+24.66}{+35.42} & +0.20 & \cit{$-$2.17}{+2.54}\\
\quad Graph (standard) $-$ no memory & +1.62 & \cit{$-$0.25}{+3.50} & +3.44 & \cit{$-$0.34}{+7.09} & +0.59 & \cit{$-$1.61}{+2.79}\\
\quad Graph (per-node) $-$ no memory, two seeds & +2.31 & \llap{\sig\hspace{1.5pt}}\cit{+0.56}{+4.06} & +8.25 & \llap{\sig\hspace{1.5pt}}\cit{+4.47}{+12.03} & $-$1.08 & \cit{$-$2.75}{+0.59}\\
\quad Text + graph (standard) $-$ no memory & +12.75 & \llap{\sig\hspace{1.5pt}}\cit{+9.75}{+15.88} & +27.49 & \llap{\sig\hspace{1.5pt}}\cit{+21.48}{+33.45} & +4.32 & \llap{\sig\hspace{1.5pt}}\cit{+1.15}{+7.60}\\
\quad Text + graph (per-node) $-$ no memory, two seeds & +12.06 & \llap{\sig\hspace{1.5pt}}\cit{+9.44}{+14.69} & +31.44 & \llap{\sig\hspace{1.5pt}}\cit{+25.95}{+36.94} & +0.98 & \cit{$-$1.67}{+3.63}\\
\midrule
\multicolumn{7}{@{}l}{\textit{Between memory routes (seed 0 unless stated)}}\\
\quad Graph: per-node (two seeds) $-$ standard & +1.25 & \cit{$-$0.94}{+3.50} & +3.95 & \cit{$-$0.69}{+8.76} & $-$0.29 & \cit{$-$2.46}{+1.87}\\
\quad Graph (per-node) $-$ text, two seeds & $-$8.75 & \llap{\sig\hspace{1.5pt}}\cit{$-$11.31}{$-$6.25} & $-$21.82 & \llap{\sig\hspace{1.5pt}}\cit{$-$27.32}{$-$16.49} & $-$1.28 & \cit{$-$3.54}{+1.08}\\
\quad Text + graph (standard) $-$ text & +1.00 & \cit{$-$1.12}{+3.12} & $-$2.06 & \cit{$-$5.21}{+1.05} & +2.75 & \cit{+0.00}{+5.59}\\
\quad Text + graph (per-node) $-$ text, two seeds & +1.00 & \cit{$-$0.38}{+2.44} & +1.37 & \cit{$-$0.52}{+3.44} & +0.79 & \cit{$-$1.08}{+2.55}\\
\quad Text + graph: per-node (two seeds) $-$ standard & $-$0.12 & \cit{$-$1.94}{+1.69} & +3.09 & \llap{\sig\hspace{1.5pt}}\cit{+0.34}{+5.84} & $-$1.96 & \cit{$-$4.52}{+0.59}\\
\midrule
\multicolumn{7}{@{}l}{\textit{Swap test, own memory $-$ swapped (per-node route)}}\\
\quad Graph model, graph swapped, step 3,000 & +1.88 & \cit{$-$0.25}{+4.00} & +7.90 & \llap{\sig\hspace{1.5pt}}\cit{+3.99}{+11.95} & $-$1.57 & \cit{$-$3.55}{+0.39}\\
\quad Graph model, graph swapped, step 6,000 & +2.12 & \cit{+0.00}{+4.25} & +6.53 & \llap{\sig\hspace{1.5pt}}\cit{+1.79}{+11.26} & $-$0.39 & \cit{$-$2.52}{+1.73}\\
\quad Text + graph model, graph swapped, step 3,000 & +0.25 & \cit{$-$1.00}{+1.50} & +1.72 & \cit{$-$0.68}{+4.23} & $-$0.59 & \cit{$-$1.97}{+0.79}\\
\quad Text + graph model, graph swapped, step 6,000 & +0.25 & \cit{$-$0.62}{+1.12} & +0.34 & \cit{$-$1.06}{+1.91} & +0.20 & \cit{$-$0.80}{+1.19}\\
\quad Text + graph model, text swapped, step 3,000 & +12.00 & \llap{\sig\hspace{1.5pt}}\cit{+9.00}{+15.00} & +31.62 & \llap{\sig\hspace{1.5pt}}\cit{+25.27}{+37.93} & +0.79 & \cit{$-$1.93}{+3.44}\\
\quad Text + graph model, text swapped, step 6,000 & +16.38 & \llap{\sig\hspace{1.5pt}}\cit{+13.25}{+19.50} & +41.92 & \llap{\sig\hspace{1.5pt}}\cit{+35.55}{+48.44} & +1.77 & \cit{$-$1.19}{+4.70}\\
\quad Text + graph model, both swapped, step 6,000 & +16.25 & \llap{\sig\hspace{1.5pt}}\cit{+13.12}{+19.38} & +41.58 & \llap{\sig\hspace{1.5pt}}\cit{+35.17}{+48.08} & +1.77 & \cit{$-$1.00}{+4.61}\\
\quad Text + graph model with swapped text $-$ no memory & $-$4.00 & \llap{\sig\hspace{1.5pt}}\cit{$-$6.88}{$-$1.12} & $-$11.00 & \llap{\sig\hspace{1.5pt}}\cit{$-$17.47}{$-$4.59} & +0.00 & \cit{$-$2.55}{+2.58}\\
\quad Graph model, graph-swap cost, step 6,000 $-$ step 3,000 & +0.25 & \cit{$-$1.88}{+2.38} & $-$1.37 & \cit{$-$4.84}{+2.04} & +1.18 & \cit{$-$1.19}{+3.53}\\
\quad Text + graph model, text-swap cost, step 6,000 $-$ step 3,000 & +4.38 & \llap{\sig\hspace{1.5pt}}\cit{+1.62}{+7.12} & +10.31 & \llap{\sig\hspace{1.5pt}}\cit{+5.28}{+15.47} & +0.98 & \cit{$-$2.05}{+4.05}\\
\midrule
\multicolumn{7}{@{}l}{\textit{Text model on perfect scene graphs, two seeds}}\\
\quad Text, perfect graphs $-$ no memory & +18.62 & \llap{\sig\hspace{1.5pt}}\cit{+15.69}{+21.56} & +38.32 & \llap{\sig\hspace{1.5pt}}\cit{+33.10}{+43.58} & +7.37 & \llap{\sig\hspace{1.5pt}}\cit{+3.95}{+10.74}\\
\quad Text: perfect $-$ our graphs & +7.56 & \llap{\sig\hspace{1.5pt}}\cit{+5.12}{+10.00} & +8.25 & \llap{\sig\hspace{1.5pt}}\cit{+4.36}{+12.20} & +7.17 & \llap{\sig\hspace{1.5pt}}\cit{+4.01}{+10.28}\\
\bottomrule
\end{tabular}
\end{table*}

\section{Implementation of the Per-Node Route}\label{app:impl}
\textbf{Design constraint.} In Qwen2.5-VL the vision encoder's features overwrite the image-token positions after the token-embedding layer, so node vectors cannot be placed in the blank image's slots by a hook on the embedding layer. We therefore add dedicated text placeholders.

\textbf{Components.} All are behind environment switches that are off by default; with them off, prompts and outputs are byte-identical to the original code.
\begin{enumerate}
\item \texttt{MOMA\_KG\_NODE\_TOKENS=N}: the answer-prompt builder (shared by training and evaluation) appends \texttt{\#\#\# Memory graph:} and $N$ copies of \tok. The token is added to the tokenizer before the LoRA wrapper is applied.
\item A \texttt{GraphNodeTokens} module computes $\mathrm{tokens}(h) = \mathrm{LayerNorm}(\mathrm{MLP}(h))\cdot s$, where $h$ is the node embedding after the existing graph projection (width 3,584), the MLP is Linear--GELU--Linear, and $s$ is a learned scalar initialised to the standard deviation of the embedding table. A learned pad vector, also scaled by $s$, fills unused placeholders, so outputs never depend on the random embedding row of the new token.
\item The existing forward hook on the token-embedding layer locates \tok\ positions from the input ids and writes the first $k = \min(\text{nodes}, N)$ node tokens there. Overflow is logged and keeps the earliest nodes (on test KGs, about 1\% of episodes for the graph-only model at $N{=}64$, maximum 66 nodes; none for the text + graph model at $N{=}80$, maximum 75).
\item \texttt{MOMA\_DISABLE\_LATE\_FUSION=1} skips the post-decoder cross-attention; \texttt{--graph-inject-layer -1} disables the pooled input route.
\item Checkpoints save and load the new module; the evaluation script rebuilds the same model from the same switches.
\end{enumerate}

\textbf{Gradient clipping.} In a smoke test the gradient norms were 871 for the pad vector, about 50 for each projector weight, and 23 for all LoRA tensors together. Training clips the global norm to 1.0, so the new module would have scaled every LoRA update down by a factor of about 40. Scaling the pad by $s$ reduced its gradient to 5. Clipping the new module's parameters separately, with the same threshold, leaves the clip for all other parameters exactly as in the other models. A 20-step training run then showed normal gradient norms (11 to 17).

\textbf{Checks before the full runs.} (i) With the switches off, answer prompts are identical to the original code. (ii) The graph changes the answer logits (maximum difference 2.4 to 2.8 between an episode's own graph and another's). (iii) Gradients reach the projector and the graph encoder, and none reach the disabled late fusion. (iv) Twenty training steps, validation and checkpointing run. (v) The evaluation script reloads the module with no missing or partial weights.

\textbf{Memory-text dropout (implemented, not evaluated).} \texttt{MOMA\_TEXT\_DROPOUT=p} blanks the memory text on a fraction $p$ of training samples (training set only; validation and test always see the full text), in the spirit of modality dropout~\cite{neverova2016moddrop}. No model has been trained with it yet; it is future work and the natural next test of the redundancy explanation. A launcher is included (\texttt{launchers/launch\_e4drop\_pernode\_v2r.sh}).

\begin{table*}[t]
\centering
\caption{Contents of \texttt{log\_corl/} and the results each file supports.}
\label{tab:logs}
\footnotesize
\begin{tabular}{@{}>{\raggedright\arraybackslash}p{0.39\textwidth}>{\raggedright\arraybackslash}p{0.39\textwidth}>{\raggedright\arraybackslash}p{0.16\textwidth}@{}}
\toprule
Path & Contents & Supports\\
\midrule
\texttt{results/e1\_e4\_test\_comparison.txt} & per-node models vs all others, with intervals; per type & Tables~\ref{tab:main}, \ref{tab:big}; Fig.~\ref{fig:headline}\\
\texttt{results/graph\_swap\_step3000.txt}, \texttt{graph\_swap\_final\_step6000.txt} & six graph/text-swap conditions, paired intervals, per type & Figs.~\ref{fig:flips}, \ref{fig:reliance}; Table~\ref{tab:swapboth}\\
\texttt{results/armc\_v2r\_test\_comparison.txt}, \texttt{armd\_v2r\_test\_comparison.txt} & graph and text + graph, standard route, vs references & Tables~\ref{tab:main}, \ref{tab:ours}\\
\texttt{test\_predictions/v2r\_kg/} & no memory and text (two seeds), graph and text + graph with both routes & Tables~\ref{tab:main}, \ref{tab:ours}; Fig.~\ref{fig:flips}\\
\texttt{training\_logs/<run>/} & full training logs: arguments, losses, gradient norms, validation accuracy and letter distribution every 500 steps, gate values & Table~\ref{tab:runs}\\
\texttt{launchers/} & exact launch scripts for every run & Table~\ref{tab:train}\\
\texttt{code/embodied\_v3\_pernode\_textdropout.diff} & per-node route, late-fusion switch, separate clip, text dropout & \secref{sec:routes}, \appref{app:impl}\\
\texttt{code/eqa\_diag\_graphswap.diff}, \texttt{diag1*.sh}, \texttt{diag1*\_analyze.py} & graph-swap control and analysis & \appref{app:swap}\\
\texttt{code/smoke\_pernode.py} & pre-launch checks (logit change, gradient flow, memory) & \appref{app:impl}\\
\texttt{code/*\_compare.py} & bootstrap comparison scripts & all tables\\
\texttt{diagnostics/graph\_swap*/} & raw outputs of the swap evaluations & Table~\ref{tab:swapboth}, Fig.~\ref{fig:flips}\\
\bottomrule
\end{tabular}
\end{table*}

\section{Hardware and Compute Requirements}\label{app:hardware}
All experiments ran on one shared server with NVIDIA H200 GPUs (140\,GiB each), 255 logical CPU cores and 2.2\,TB of RAM. Each training run and each test evaluation used a single GPU. Table~\ref{tab:hardware} gives the time of every stage. Training times run from the start time in each training log to the file time of the step-6,000 checkpoint.

\begin{table}[t]
\centering
\caption{Hardware and wall-clock time per stage. Fine-tuning covers the runs of Table~\ref{tab:runs}.}
\label{tab:hardware}
\footnotesize
\setlength{\tabcolsep}{3pt}
\begin{tabular}{@{}>{\raggedright\arraybackslash}p{0.37\columnwidth}c>{\raggedright\arraybackslash}p{0.49\columnwidth}@{}}
\toprule
Stage & GPUs & Wall-clock time\\
\midrule
Re-render frames at 512\,px (AI2-THOR, Vulkan) & 1 & 7.4\,min for the 8,335 test frames\\
Scene graphs (MomaGraph-R1, 512\,px) & 2 & train (206,899 frames) 17.8\,h; validation 2.3\,h; test 45\,min; about 0.6\,s per frame per GPU\\
Fine-tune, MomaGraph-R1 & 1 & about 17--30\,h per run\\
Test evaluation (800 questions) & 1 & 23--36\,min\\
\bottomrule
\end{tabular}
\end{table}

\textbf{Totals.} Each training run took about 17 to 30 GPU-hours, and building the scene graphs for all three splits about 40 GPU-hours.

\textbf{Memory and storage.} The MomaGraph-R1 weights take 16.8\,GB in bf16. Peak GPU memory was not logged. Two runs often shared one card (the two per-node models trained on the same GPU at the same time), so two fine-tunes fit together in 140\,GiB. A saved checkpoint holds the LoRA weights and any graph modules: 0.4\,GB for no memory and text, 0.7\,GB for the per-node route, 1.0\,GB for the standard graph route.

\textbf{Variation.} Identical settings differ by up to $1.8\times$ in time because other jobs shared the server. The paired bootstrap, tables and charts need no GPU; they are recomputed from the released logs on a laptop CPU.

\section{Logs and Artifacts}\label{app:logs}
\begin{table*}[t]
\centering
\caption{Training runs reported in this paper, in the order they started. Validation: answer accuracy (\%) on 400 validation questions at the
  last validation pass and at the best one; letter: largest share of one answer letter at any validation pass. Gate: final late-fusion $\gate$
  (standard graph route only). Test: correct of 800 at the final checkpoint.}
\label{tab:runs}
\setlength{\tabcolsep}{3.6pt}
\footnotesize
\renewcommand{\arraystretch}{0.96}
\begin{tabular}{@{}llccccccl@{}}
\toprule
Memory & Graph route & Seed & Steps & Val.\ last & Val.\ best (step) & Letter & Gate & Test\\
\midrule
\multicolumn{9}{@{}l}{\textit{No memory (reads no scene graph)}}\\
No memory & -- & 0 & 6{,}000 & 67.3 & 69.8 (5k) & 0.47 & -- & 514\\
No memory & -- & 1 & 6{,}000 & 67.8 & 68.3 (2.5k) & 0.46 & -- & 523\\
\midrule
\multicolumn{9}{@{}l}{\textit{Our scene graphs}}\\
Text & -- & 0 & 6{,}000 & 79.0 & 80.3 (5.5k) & 0.44 & -- & 608\\
Text & -- & 1 & 6{,}000 & 80.0 & 80.0 (6k) & 0.41 & -- & 606\\
Text + graph & standard & 0 & 6{,}000 & 76.8 & 77.3 (4k) & 0.41 & 0.013 & 616\\
Graph & standard & 0 & 6{,}000 & 68.5 & 69.8 (3.5k) & 0.54 & 0.062 & 527\\
Graph & per-node & 0 & 6{,}000 & 71.5 & 72.3 (5.5k) & 0.48 & -- & 532\\
Text + graph & per-node & 0 & 6{,}000 & 78.5 & 78.5 (6k) & 0.40 & -- & 610\\
Graph & per-node & 1 & 6{,}000 & 72.3 & 73.3 (5.5k) & 0.55 & -- & 542\\
Text + graph & per-node & 1 & 6{,}000 & 77.5 & 78.0 (3k) & 0.43 & -- & 620\\
\bottomrule
\end{tabular}
\end{table*}

Every number in this paper can be recomputed from the \texttt{log\_corl/} folder released with it (also as a single zip). For review it will be provided anonymised. Run names in the logs: \texttt{nokg} = no memory, \texttt{memtext} = text, \texttt{kg} = graph, \texttt{kgmem} = text + graph; \texttt{\_sgv2r} = the scene graphs of \secref{sec:sg}; \texttt{e1\_pernode}, \texttt{e4\_pernode} = graph and text + graph with the per-node route. Result files call the four memory types arms A to D and the two per-node models E1 and E4.

\textbf{Perfect scene graphs.} The two text runs on perfect scene graphs come from a separate log bundle of the same codebase, whose no-memory runs are the same runs as ours (identical per-question outputs). Their per-question predictions are added to \texttt{test\_predictions/}; they are not in the counts of Tables~\ref{tab:hardware} and~\ref{tab:runs}.

\textbf{Run log.} Table~\ref{tab:runs} lists the training runs reported in this paper; all completed 6,000 optimizer steps with no skipped step.

\textbf{Order of the experiments.} The no-memory runs ran on 15 and 17 Sep. The text, graph and text + graph models on our scene graphs followed (1 to 6 Oct). The two per-node models started on 7 Oct, after the standard graph model had finished, and each was trained and tested once; their node budgets (64, 80) were set from KG node counts before launch. The swap test was run first on the step-3,000 snapshots and then, on 8 Oct, on the final checkpoints. Second seeds of both per-node models ran from 18 to 19 Oct with identical settings; each was trained and tested once.

\section{Frequently Asked Questions}\label{app:faq}

\textbf{Is the per-node graph model's gain more than the effect of extra prompt tokens?} Yes. The placeholders are in the prompt whether the graph is right or wrong, so the graph-swap test isolates content. With another episode's graph, the model loses 6.53\sig\ points on route questions and its rooms-visited accuracy falls from 66.4\% to 47.4\% (\secref{sec:swap}).

\textbf{Does the per-node text + graph model's tie with text mean the graph is useless?} It means the model does not use the graph when the text is present. Swapping its graph changes 12 answers, as many as re-evaluating the same checkpoint, while swapping its text costs 16.38\sig\ points. The same per-node route does carry information in the graph-only model, which has no text.

\textbf{Why only one seed for the standard graph and text + graph models?} The per-node models have two seeds; each fine-tune takes about a day on one GPU. Our intervals resample episodes, so they cover question sampling but not seed variation. Where two seeds exist, runs of the same model differ by 1 to 10 correct answers (Table~\ref{tab:runs}).

\textbf{Are differences of one or two points meaningful?} Usually not. Re-evaluating an unchanged checkpoint changes 9 to 12 answers, and two seeds of the same model differ by 1 to 10 correct answers in total. We call a difference significant only when its paired interval excludes zero, and we compare conditions within one evaluation session where possible (\appref{app:swap}).

\textbf{Why blank the image at answer time?} So that every episodic fact must come from memory. The no-memory model then measures what the question and options alone support (64.2\% and 65.4\%).

\textbf{Why was no graph model trained on perfect scene graphs?} Each run takes about a day, and we spent the runs on the per-node route and the swap test. A graph-only model on perfect graphs would show whether the per-node route's limit is the route or the graph content; it is the most direct next run.

\textbf{Why not simply use text memory?} Text is the strongest route here and the baseline any graph route must beat. A graph route earns its cost only if it adds what text cannot express, such as 3D positions or the physical rooms of objects (\secref{sec:conclusion}).

\textbf{Would another graph swap give a different cost?} Possibly. We used one fixed derangement (\appref{app:swap}), so the cost is an average over 200 donor pairings for one assignment. Repeating the test over several derangements and reporting the spread is the next check; the text + graph model's text-swap cost (+16.38\ci{+13.25}{+19.50}) is far from zero, while the graph-only model's route-question cost (+6.53\ci{+1.79}{+11.26}) is closer.

\textbf{Are the intervals corrected for multiple comparisons and for houses?} No. They are unadjusted episode-bootstrap intervals. \appref{app:stats} names the primary comparisons, marks the rest as exploratory, and lists the results that a house-level bootstrap or a multiplicity correction could change.

\textbf{Is the per-node route the contribution?} Only in part. Its accuracy gain over the standard route is small and within noise (+1.25\ci{$-$0.94}{+3.50}), and with text present the two routes tie. Its use here is that its graph-only model depends on the graph, which lets the swap test be checked on a case where dependence exists. Text memory is the reference, not a proposed method.

\textbf{Does a fixed number of node tokens drop nodes?} Rarely. The graph-only model's 64 slots fall short only for KGs of 65 or 66 nodes, and the text + graph model's 80 slots exceed its largest KG (75 nodes); on overflow the earliest nodes are kept (\appref{app:impl}).

\textbf{Why score the final checkpoint and not the best validation one?} So that no run benefits from checkpoint selection. In every run the final validation accuracy is within 5 points of the best pass (Table~\ref{tab:runs}).

\textbf{Was the test set used to design the per-node route?} No. The design follows from the standard route's failure to deliver graph information in earlier development runs. The two per-node models were each trained and evaluated once, with node budgets fixed from KG sizes at launch; the test set was used again only for the swap analysis.

\end{document}